\documentclass[10pt,twocolumn,letterpaper]{article}

\usepackage[final]{cvpr} 

\usepackage{graphicx}
\usepackage{amsmath}
\usepackage{amssymb}
\usepackage{booktabs}

\usepackage{paralist}
\usepackage{multirow}
\usepackage[accsupp]{axessibility}  

\usepackage[pagebackref,breaklinks,colorlinks,bookmarks=false]{hyperref}

\DeclareGraphicsExtensions{.pdf,.jpg,.png}
\graphicspath{{figs/}}

\newcommand{\secref}[1]{Section~\ref{sec:#1}}
\newcommand{\figref}[1]{Figure~\ref{fig:#1}}
\newcommand{\tabref}[1]{Table~\ref{tbl:#1}}

\def\cvprPaperID{990} 
\def\confName{CVPR}
\def\confYear{2022}

\begin{document}

\title{Revisiting Near/Remote Sensing with Geospatial Attention}

\author{
  \centering
  \begin{minipage}{.9\linewidth}
    \centering
    \begin{minipage}{1.4in}
      \centering
      Scott Workman
      \\[.15cm]
      \normalsize{DZYNE Technologies}
    \end{minipage}
    \begin{minipage}{1.4in}
      \centering
      M. Usman Rafique
      \\[.15cm]
      \normalsize{Kitware, Inc.}
    \end{minipage}
    \begin{minipage}{1.4in}
      \centering
      Hunter Blanton
      \\[.15cm]
      \normalsize{University of Kentucky}
    \end{minipage}
    \begin{minipage}{1.4in}
      \centering
      Nathan Jacobs
     \\[.15cm]
      \normalsize{University of Kentucky}
    \end{minipage}
  \end{minipage}
}

\maketitle

\begin{abstract}

    This work addresses the task of overhead image segmentation when auxiliary ground-level images are available. Recent work has shown that performing joint inference over these two modalities, often called near/remote sensing, can yield significant accuracy improvements. Extending this line of work, we introduce the concept of geospatial attention, a geometry-aware attention mechanism that explicitly considers the geospatial relationship between the pixels in a ground-level image and a geographic location. We propose an approach for computing geospatial attention that incorporates geometric features and the appearance of the overhead and ground-level imagery. We introduce a novel architecture for near/remote sensing that is based on geospatial attention and demonstrate its use for five segmentation tasks. The results demonstrate that our method significantly outperforms the previous state-of-the-art methods.

\end{abstract}

\section{Introduction}

Accurately monitoring the Earth's surface is critical to many scientific fields and to society at large. Important applications include weather forecasting, disaster response, population density estimation, and environmental monitoring. Traditionally these applications have relied on {\em remote sensing} approaches applied to overhead imagery from satellite or airborne cameras. Computer vision techniques have long been applied to such imagery to automate various tasks~\cite{van1988knowledge,healey1996using,pan2005band}, including recent work on detecting roads~\cite{mattyus2017deeproadmapper}, estimating land cover~\cite{robinson2019large}, understanding traffic flow~\cite{workman2020dynamic}, and constructing dynamic visual attribute maps~\cite{salem2020learning}.

In addition, the use of imagery from alternative sources, such as consumer devices~\cite{wang2013observing} and webcams~\cite{jacobs2009global}, has been explored for various monitoring applications. For example, geotagged ground-level images, including consumer photographs, have been used to monitor weather~\cite{wang2013observing}, estimate geo-informative attributes~\cite{lee2014predicting}, and characterize safety~\cite{arietta2014city}. Similarly, webcam imagery has been used for vegetation~\cite{sonnentag2012digital}, snow cover~\cite{portenier2020towards}, and marine debris~\cite{kako2018sequential} monitoring. This class of methods, often referred to as {\em proximate sensing}~\cite{leung2010proximate} or {\em image-driven mapping}, uses large georeferenced photo collections to derive geospatial information.

\begin{figure}
    \centering
    
    \setlength{\fboxsep}{0pt}
    \setlength{\fboxrule}{1pt}
    
    \begin{subfigure}{.3855\linewidth}
        \centering
        \fbox{\includegraphics[width=.976\linewidth,trim=1cm 1cm 1cm 1cm, clip]{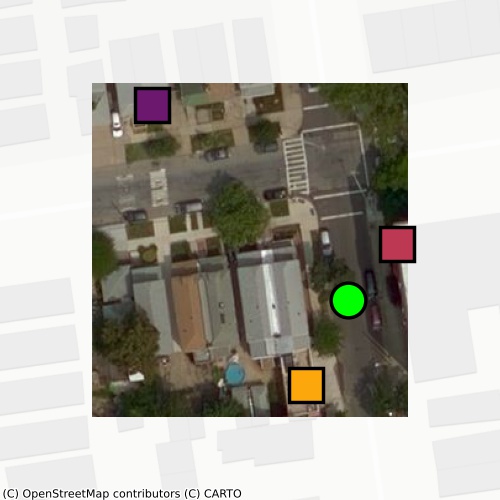}}
        \includegraphics[width=1\linewidth]{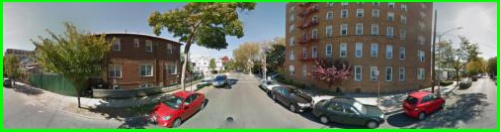}
    \end{subfigure}
    \hfill
    \begin{subfigure}{.584\linewidth}
        \centering
        \includegraphics[width=1\linewidth]{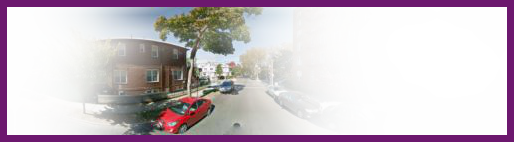}
        \includegraphics[width=1\linewidth]{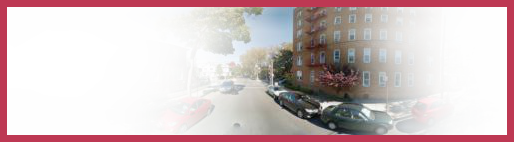}
        \includegraphics[width=1\linewidth]{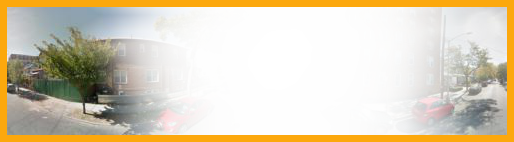}
    \end{subfigure}
    
    \caption{We introduce a novel neural network architecture that uses geospatial attention in the setting of near/remote sensing. Our approach operates on an overhead image and a set of nearby ground-level panoramas, enabling optimal feature extraction for a query location (square) from each ground-level image (circle) in a manner that is ``geometry-aware''.}

    \label{fig:cartoon}
\end{figure}

These two strategies, remote and proximate sensing, offer complementary viewpoints of the world. Overhead imagery is widely available at increasingly high resolutions and has dense coverage. However, fine-grained properties are often hard to characterize from only a top-down perspective~\cite{workman2017natural}. In contrast, geotagged ground-level images are sparsely distributed but capture high resolution, semantically rich details. To convert these sparse samples into a dense map, an additional process, such as geometric warping or locally weighted averaging, is required. This means that using only ground-level imagery results in either missing values for areas that are not imaged or low spatial resolution outputs~\cite{arietta2014city}.  

Combining these two modalities, which we refer to as {\em near/remote sensing}, has emerged as a compelling research area that addresses weaknesses in methods that only use a single modality. Early techniques focused on building explicit geometric models~\cite{frueh2003constructing}. Our work is more closely related to methods that attempt to extract semantic information, such as that of Luo et al.~\cite{luo2008event} on event recognition. Other methods have been proposed that consider how to relate information from pairs of co-located ground-level and overhead images~\cite{lin2013cross,workman2015wide,workman2021augmenting}. Recently, network architectures have been proposed that allow for combining an overhead image with nearby ground-level images for general segmentation tasks~\cite{workman2017unified,cao2018integrating}.

The standard approach is to extract image features from nearby ground-level images, fuse them to form a dense grid of features that is geospatially aligned with features extracted from the overhead image, and concatenate the two feature sets for joint inference. Though this strategy has shown great promise versus single-modality alternatives, there remains significant room for improvement. One major limitation of current approaches is the use of global image features, which ignore important geometric information. A new approach is needed in order to extract meaningful geo-informative features from each sample for the given task.

In this work we introduce the concept of geospatial attention. As opposed to a standard spatial attention module (e.g., \cite{woo2018cbam}), which operates solely on an input feature map to identify salient regions, geospatial attention additionally considers the geospatial relationship between the input and a target location, with the goal of identifying meaningful geo-informative regions. The key insight is that for many tasks, the position and orientation of the input relative to a location of interest is crucial for optimally fusing information from multiple sources (\figref{cartoon}). We propose a method for estimating geospatial attention that incorporates the semantic content of the input image in addition to geometry and overhead appearance, with the goal of identifying geo-informative regions of the input. 

We introduce a novel neural network architecture that uses geospatial attention in the setting of near/remote sensing. Our approach operates on an overhead image and a set of nearby ground-level panoramas. It simultaneously learns to extract features from each image modality in an end-to-end fashion. To support evaluation, we extend an existing dataset with two new per-pixel labeling tasks. Extensive evaluation demonstrates the utility of our approach for five labeling tasks: land use, building age, building function, land cover, and height. Significant improvements in accuracy are observed relative to previous work and an internal ablation study is used to highlight the most important components.

\begin{figure*}
  \centering
  \includegraphics[width=1\linewidth]{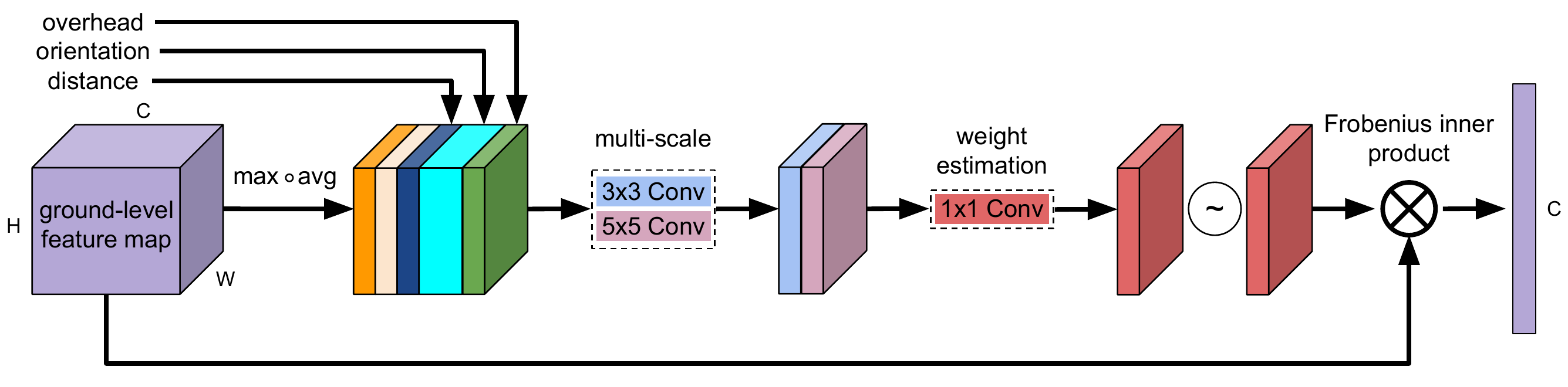}
  \caption{Our geospatial attention architecture, which we use to reduce a feature map to a geo-informative feature vector.}
  \label{fig:geo-attention}
\end{figure*}

\section{Related Work}

Numerous papers have explored the use of neural network architectures for overhead image segmentation, including survey papers in remote sensing venues~\cite{zhang2016deep,ma2019deep} and papers addressing specific tasks at computer vision venues~\cite{azimi2019skyscapes,mattyus2017deeproadmapper,Hamaguchi_2018_CVPR_Workshops,robinson2019large}. Given this, we focus on work using ground-level images for mapping applications,  fusing ground-level and overhead imagery, and attention mechanisms.

\subsection{Image Driven Mapping}

The availability of large numbers of geotagged ground-level images, from social media and camera-equipped vehicles, has made it possible to construct maps of various visual phenomena. Numerous papers have explored this approach; we only highlight a few. 
Crandall et al.~\cite{crandall2009mapping} was one of the first works to highlight the potential of geotagged social media imagery for understanding locations. Zhou et al.~\cite{zhou2014recognizing} and Arietta et al.~\cite{arietta2014city} both propose to use such imagery for understanding urban areas. Similarly, Leung and Newsam~\cite{leung2015land} address the task of (coarse) land-cover classification using ground-level images.

The main limitation of approaches that rely exclusively on ground-level imagery is that they are not able to generate high-resolution output maps, especially when mapping areas away from major tourist destinations. Several works~\cite{zhu2015land,zhu2019fine,srivastava2019understanding} have addressed this using building footprints from GIS data. However, these approaches are generally limited to object-level classification and are thus unsuitable for many tasks. Our approach, in contrast, does not require GIS vector data and uses a more general strategy for geometric fusion.

\subsection{Mapping Using Ground-Level and Overhead Imagery}

Techniques for combining ground-level and overhead imagery hold significant promise for improving our ability to construct fine-grained, accurate maps. Lef\'evre et al.~\cite{seamlessMultiview} provide an early overview of this promise, including extensions to earlier work on tree detection and classification~\cite{wegner2016cataloging}. M{\'a}ttyus et al.~\cite{mattyus2016hd} address the task of roadway mapping. The first work to address the general near/remote segmentation task~\cite{workman2017unified} used a kernel weighted averaging approach for fusing ground-level feature vectors. We improve upon this by additionally performing adaptive pooling of the ground-level image features. Hoffmann et al.~\cite{hoffmann2019model} evaluate feature fusion and decision fusion approaches for coarse classification tasks, but rely on a single ground-level image oriented toward the building. Our work can be seen as extending this approach from classification to dense segmentation, from single to multiple ground-level images, and from hard to soft attention.

\subsection{Cross-view Localization and Synthesis}

Closely related to the near/remote segmentation task are the tasks of localizing a ground-level image using overhead reference imagery and predicting the visual appearance of a ground-level image from an overhead image. Both benefit from reasoning about the geometric relationship between ground-level and overhead views.

Early work on cross-view geolocalization focused on arbitrarily oriented perspective images~\cite{lin2013cross,lin2015learning,workman2015geocnn,workman2015wide,vo2016localizing}, but more recent methods have emphasized localization of north-aligned on-street panoramas. The top performing approaches for this task now explicitly integrate the geometric relationship between the ground-level and overhead perspectives. Liu and Li~\cite{liu2019lending} add an orientation map to each input image. Several papers have incorporated spatial attention mechanisms: Cai et al.~\cite{cai2019ground} use a multi-scale variant of CBAM~\cite{woo2018cbam} and Shi et al.~\cite{shi2019spatial} show attention helps when the input domains are aligned. We use these approaches as building blocks for implementing geospatial attention.

In cross-view synthesis, a key challenge is predicting the geometric transformation. Early work by Zhai et al.~\cite{zhai2017predicting} used an implicit model, and more recent work~\cite{lu2020geometry,toker2021coming,shi2022geometry} has shown that more explicit geometric models can lead to improved results.

\subsection{Attention Mechanisms}

Various attention mechanisms have been introduced, including spatial transformer networks~\cite{jaderberg2015spatial}, which apply hard attention based on a parametric transformation, and methods that use learning-based soft attention~\cite{jetley2018learn}, channel-wise attention~\cite{chen2017sca}, and self-attention~\cite{parmar2018image}. Recent work in cross-view matching has introduced geometric attention models~\cite{cai2019ground,shi2019spatial}, but they focus on a single known transformation. He et al.~\cite{he2020epipolar} introduce an attention model which is similar to self-attention but incorporates epipolar geometry. They demonstrate its use for human pose estimation~\cite{He_2020_CVPR_Workshops}. Our approach is focused on learning to predict attention in a ground-level image relative to a geographic location, using both geometric and image features to inform the weighting.

\begin{figure*}
    \centering
    \includegraphics[width=1\linewidth]{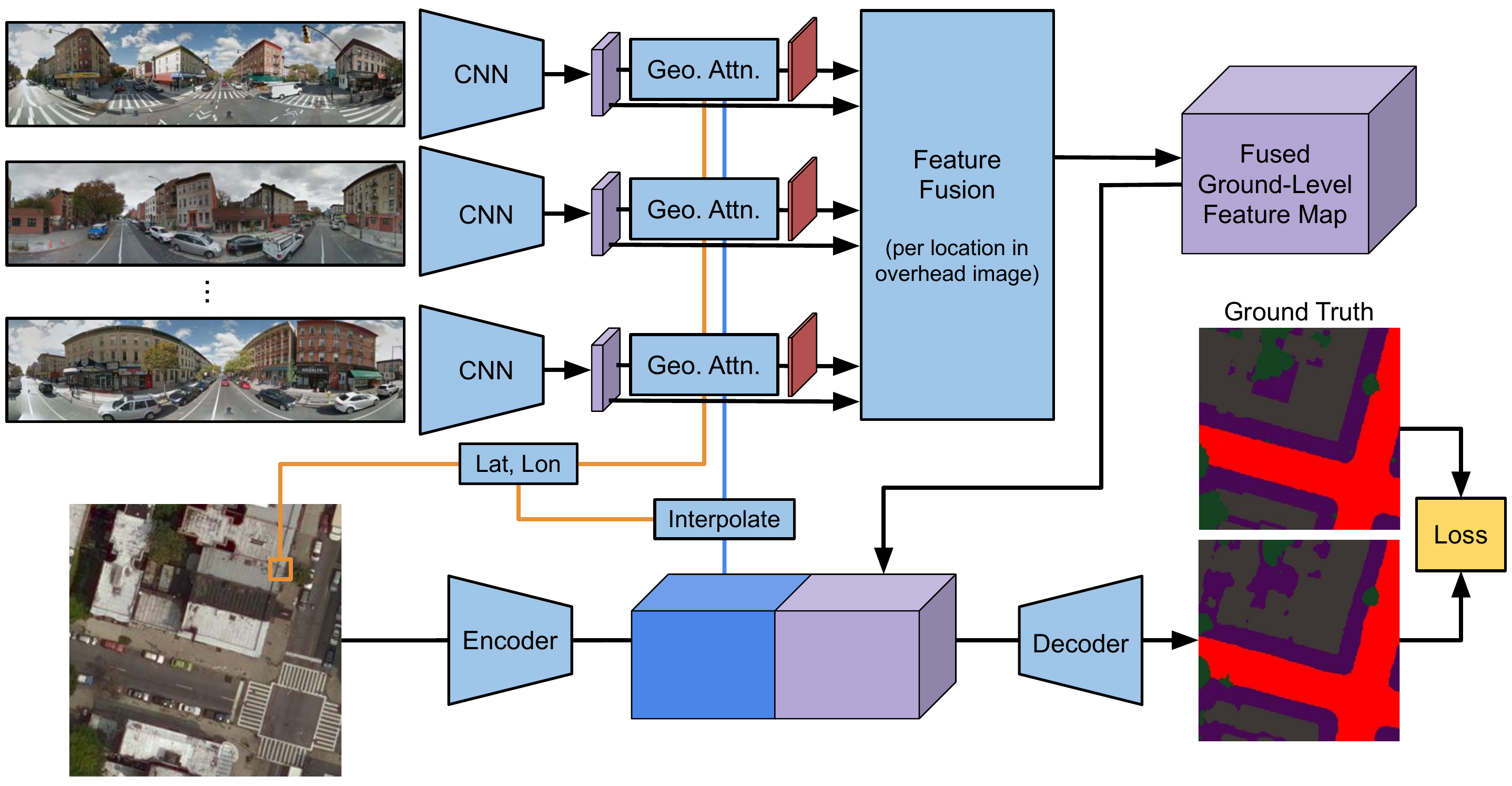}
    \caption{An overview of our architecture for near/remote sensing.}
    \label{fig:architecture}
\end{figure*}

\section{Geospatial Attention}

We address the task of overhead image segmentation, which includes semantic tasks, such as land use classification, and geometric tasks, such as building height estimation. We assume we are given a geo-registered overhead image and, for simplicity, that we output the segmentation as a pixel-level labeling. In addition to the overhead image, we are given a set of nearby ground-level images with known intrinsic and extrinsic calibration (georeferenced). The key challenge is to combine information from all available images to make an accurate segmentation map. To address this challenge, we propose {\em geospatial attention}, a geometry-aware attention mechanism, and a neural-network architecture that uses geospatial attention to fuse information from the ground-level images.

\subsection{Defining Geospatial Attention}

Like previous spatial attention mechanisms, the objective of geospatial attention is to identify important regions of an input image, or equivalently an input feature map, by estimating an attention value for each image pixel. Geospatial attention extends this idea by parameterizing the attention map by a target geographic location. Since the image is fully calibrated, we know the camera center location and have a ray, in scene coordinates, associated with each pixel. Together, geospatial attention will depend on the distance between the target location and the camera location, the target-relative orientation of each pixel ray, and image features.

We use geospatial attention to reduce a feature map into a feature vector as follows. Given an input feature map, $F_i \in \mathbb{R}^{H \times W \times C}$, extracted from a ground-level image, $I_i$, at camera location, $l_i$, and a target location, $l_t$, geospatial attention infers a 2D spatial attention map $P_{i,t} \in \mathbb{R}^{{H \times W}}$. This process is visualized in \figref{geo-attention}. Similar to Shi et al.~\cite{shi2019spatial}, we use the attention map, $P_{i,t}$, to generate a feature vector, $K_i = \{k^c\}, c=1,...,C$, as:
\begin{equation}
    k^c = \langle \mathbf{f}^c, P_{i,t} \rangle_F
\end{equation}
where $\mathbf{f^c} \in  \mathbb{R}^{H \times W}$ represents the c-th channel of $F_i$, $\langle.,.\rangle_F$ denotes the Frobenius inner product, and $k^c$ is the feature output for the c-th channel. The resulting feature vector represents the information from the input feature map, $F_i$, that is relevant to the target location, $l_t$. 

\subsection{Inferring Geospatial Attention}

The key component of geospatial attention is the computation of the spatial attention map, $P_{i,t}$, from the input feature map, $F_i$, camera location, $l_i$, and target location, $l_t$. We represent this computation as a convolutional network applied to an augmented version of the input feature map, which is constructed as follows. 

From the input and target locations, ($l_i$, $l_t$), we calculate two geometric feature maps. The first is the haversine distance $d$ (meters) between $l_i$ and $l_t$, which is tiled to $H \times W$. The second is the orientation $\theta$ of each pixel, relative to the target location, $l_t$, which is represented as a $H \times W \times 3$ tensor. This is computed by rotating the original pixel rays, which are initially in an east-north-up coordinate frame, about the zenith direction so that $[0,1,0]$ points toward the target location. We also construct two image feature maps. First, the feature vector from the overhead image at the target location, $S(l_t)$, is tiled to $H \times W \times N$. We then apply max-pooling and average-pooling operations along the channels dimension to both the input feature map, $F_i$, and the tiled overhead feature map, resulting in two $H\times W \times 2$ feature maps.

The geometric feature maps and the pooled image feature maps are concatenated to produce a $H \times W \times 8$ tensor which we use as input to a convolutional network for inferring spatial attention. First, two convolutional layers, $3\times3$ and $5\times5$ respectively, are used to generate intermediate feature maps, similarly to Cai et al.~\cite{cai2019ground}. These intermediate feature maps are concatenated and passed to a $1 \times 1$ convolution, with a sigmoid activation, to estimate the spatial attention map, $P_{i,t}$.

\section{An Architecture for Near/Remote Sensing}

We propose a high-level neural network architecture for the task of near/remote sensing. Our architecture, visualized in \figref{architecture}, has three primary components. First, we extract features from each image modality (\secref{image_feature}). Next, we use geospatial attention to generate a spatially consistent, dense grid of geo-informative features from the set of nearby ground-level images (\secref{grid}). Finally, we fuse the dense ground-level feature map with the overhead image feature map and use that as input to a decoder that generates the segmentation output (\secref{function}). All components are differentiable, enabling end-to-end optimization of the low-level feature extraction networks and the attention model for the given segmentation task. The remainder of this section describes the high-level architecture, see the supplemental materials for additional details.

\subsection{Feature Encoders}
\label{sec:image_feature}

For each input image, we use a convolutional feature encoder to extract a feature map. Each output location in the feature map has a feature vector and a geometric descriptor, which depends on the modality. While we present specific choices for the feature encoders, we emphasize that there are many alternatives that could be used to achieve application requirements (e.g., less memory, lower computational complexity, or, potentially, higher accuracy).

\paragraph{Overhead Image} 

To extract feature maps from the overhead image, we use an EfficientNet-B4~\cite{tan2019efficientnet} encoder. We use the output activations from the second and third stages. Given an input image of size $256 \times 256$, the resulting feature maps are $64 \times 64 \times 32$ and $32 \times 32 \times 56$. We also calculate the geographic location of each element in the final feature map, resulting in a $32 \times 32 \times 2$ geolocation map. These will be used as target locations for computing geospatial attention.

\paragraph{Ground-level Images}

To extract a feature map from the ground-level images, we use ResNet-50~\cite{he2016deep} pretrained on ImageNet. We use the output activations from the fourth stage. In our case, we operate directly on panoramas in an equirectangular projection, and we crop the top and bottom, approximately $40^\circ$ from both, to minimize distortion. After cropping and resizing the panoramas, the input image size is $128 \times 500$. The resulting feature map for each image is $8 \times 32 \times 1024$. We add a $1 \times 1$ convolution, with LayerNorm~\cite{ba2016layer} and a ReLU activation, to reduce the number of channels to $128$. Each column in the ground-level feature map is associated with a geographic direction, with the center column initially corresponding to north. To facilitate geospatial attention, we compute the pixel rays for each image feature location. We also record the location of the camera center, which is the same for all features.

\subsection{Fusing Ground-Level Image Features}
\label{sec:grid}

We use geospatial attention to construct a spatially consistent, dense grid of geo-informative features. This process combines features from the set of nearby ground-level images, using the $32\times 32$ grid of geolocations from the overhead image as the target locations. In the remainder of this section, we describe the method for computing the feature vector associated with a single target location.

For a given target location, we first apply geospatial attention to reduce the individual ground-level image feature maps to feature vectors. In addition, we record the sum of the spatial attention map for each image (i.e., a scalar representing total attention). We then perform a weighted average to combine features from all ground-level images. The weight for this operation is computed by applying a softmax, across all ground-level images, to the recorded total attention values. This process is repeated, in parallel, for each target location, and the resulting vectors are concatenated to form a dense grid.

The result is a feature map that represents the same geographic extent as the overhead image feature map, but is based on features extracted from the ground-level images. Given that the overhead feature map is $32 \times 32$ and each ground-level image feature has $128$ dimensions, the resulting dense grid is $32 \times 32 \times 128$.

\subsection{Segmentation Decoder}
\label{sec:function}

The final component of our architecture is a decoder that generates the segmentation output. We use a U-Net~\cite{ronneberger2015u}-style decoder, which expects four input feature maps. For the first two, we use the feature maps extracted from the overhead image, with spatial size $64\times 64$ and $32\times 32$ respectively. For the last two, we first concatenate features extracted from the overhead imagery and the dense grid of fused-features from the ground-level images, obtaining a $32 \times 32 \times 184$ feature map. This is passed through two blocks, each with three convolutional layers (BatchNorm~\cite{ioffe2015batch}, ReLU), to create two additional feature maps of size $16 \times 16 \times 160$ and $8 \times 8 \times 448$ respectively. These are used as the last two inputs to the decoder. The decoder has five upsampling blocks with skip connections, the output of which is then passed through a final convolutional layer with an activation and number of channels that depends on the target label.

\subsection{Implementation Details}

We implement our methods using Pytorch~\cite{paszke2019pytorch} and Pytorch Lightning~\cite{falcon2019pytorch}. Our networks are optimized using Adam~\cite{kingma2014adam} with the initial learning rate set to $1e^{-4}$. All networks are trained for 25 epochs with a learning rate policy that decays the learning rate by $\gamma=0.96$ every epoch. For classification tasks, we use the cross-entropy loss. For regression tasks, we use the uncertainty loss from Kendall and Gal~\cite{kendall2017uncertainties}. 

\begin{table*}
  \centering
  \caption{Brooklyn evaluation results.}
  
  \resizebox{\linewidth}{!}{
      \begin{tabular}{@{}lccccccrccc@{}}
        \toprule
        & \multicolumn{2}{c}{Land Use} & \multicolumn{2}{c}{Age} & \multicolumn{2}{c}{Function} & \multicolumn{2}{c}{Land Cover} & \multicolumn{2}{c}{Height} \\
        & \multicolumn{1}{c}{mIOU} & \multicolumn{1}{c}{Acc} & \multicolumn{1}{c}{mIOU} & \multicolumn{1}{c}{Acc} & \multicolumn{1}{c}{mIOU} & \multicolumn{1}{c}{Acc} & \multicolumn{1}{c}{mIOU} & \multicolumn{1}{c}{Acc} & \multicolumn{1}{c}{RMSE} & \multicolumn{1}{c}{RMSE log} \\
        \bottomrule
        {\em Workman et al.~\cite{workman2017unified}} & 45.54\% & 77.40\% & 23.13\% & 43.85\% & 14.59\% & 44.88\%   & & & & \\
        {\em Cao et al.~\cite{cao2018integrating}}     & 48.15\% & 78.10\% & & & & & & & & \\
        \hline
        {\em proximate} & 49.82\% & 75.30\%  & 36.68\% & 56.48\% & 12.13\% & 43.81\% & 38.27\% & 67.63\% & 4.440 & 1.031 \\
        {\em remote} & 40.30\% & 72.98\% & 16.40\% & 34.43\% & 4.50\% & 34.53\%& 69.48\% & 86.71\%   & 3.260 & 0.785 \\
        {\em ours} & \textbf{69.24\%} & \textbf{86.82\%} & \textbf{51.70\%} & \textbf{70.34\%} & \textbf{27.40\%} & \textbf{60.31\%} & \textbf{74.59\%} & \textbf{88.10\%} & \textbf{2.845} & \textbf{0.747} \\
        \bottomrule
      \end{tabular}
  }
  \label{tbl:brooklyn_results}
\end{table*}

\section{Experiments}

We evaluate our approach on five labeling tasks and find that our method significantly outperforms the previous state-of-the-art methods.

\subsection{Datasets}

For this work, we use the Brooklyn and Queens dataset~\cite{workman2017unified}. The dataset contains non-overlapping overhead images (approx.\ 30~cm resolution) for New York City as well as a large reference database of ground-level panoramas collected from Google Street View.  We define two new per-pixel labeling tasks, estimating land cover and estimating height. For height, we use a highest-hit digital surface model (DSM) derived from topographic and bathymetric LiDAR data collected for New York City in 2017 (approx.\ 30~cm resolution). For land cover we use a \mbox{6\thinspace in} resolution, 8-class land cover dataset that was similarly derived from the 2017 LiDAR capture. This data is made publicly available through the NYC OpenData portal. We relate this data to the overhead images in the Brooklyn and Queens data to generate a ground-truth per-pixel labeling. Including our two new tasks, there are five tasks for this dataset: estimating land use, building age, building function, land cover, and height. In all experiments, we include the 20 closest street-level panoramas to each overhead image.

\subsection{Baselines}

For evaluating our proposed architecture, we consider several baseline methods that use components of our full approach:
\begin{compactitem}
    \item \emph{remote}: a traditional remote sensing approach that only uses overhead imagery. We start from our full approach but omit the ground-level feature map.
    \item \emph{proximate}: a proximate sensing approach that only uses geotagged ground-level imagery. We start from the ground-level feature maps, use geospatial attention (without overhead image features) to construct a dense feature grid, and then use a similar U-Net~\cite{ronneberger2015u} style decoder, without overhead image features and the associated skip connections, to generate the output.
\end{compactitem}
Additionally, we compare against prior results from Workman et al.~\cite{workman2017unified} and Cao et al.~\cite{cao2018integrating}. Both of these methods follow a similar strategy in using locally weighted averaging to construct the dense ground-level feature map, with additional differences in the choice of ground-level feature extractor and segmentation architecture.

\subsection{Quantitative Results}

For classification tasks, we report results using pixel accuracy and region intersection over union averaged over classes (mIOU). For both of these metrics, higher is better. For regression tasks, we report results using root mean square error (RMSE) and root mean square log error (RMSE log). As in previous work~\cite{workman2017unified}, when computing these metrics we ignore any ground-truth pixel labeled as unknown. Further, for the building age and function estimation tasks, we ignore pixels labeled as background.

\begin{figure*}

  \centering
  
  \setlength\tabcolsep{1pt}
  \newcommand\w{.12\linewidth}    
    
    \begin{tabular}{cccccccc}
    
      \multicolumn{2}{c}{Height} & \multicolumn{2}{c}{Land Cover} & \multicolumn{2}{c}{Age} & \multicolumn{2}{c}{Land Use} \\
    
      \includegraphics[width=\w]{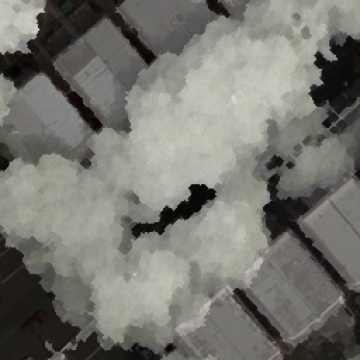} &
      \includegraphics[width=\w]{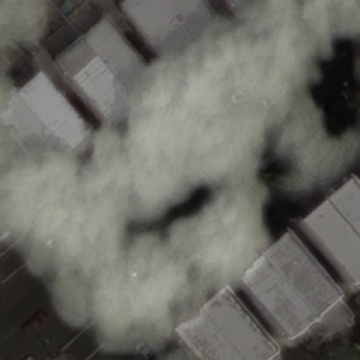} &
      \includegraphics[width=\w]{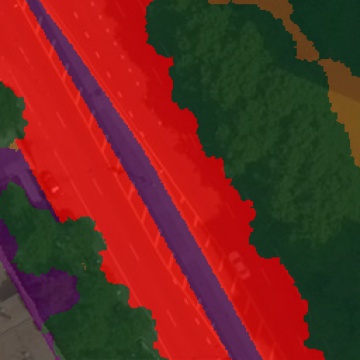} &
      \includegraphics[width=\w]{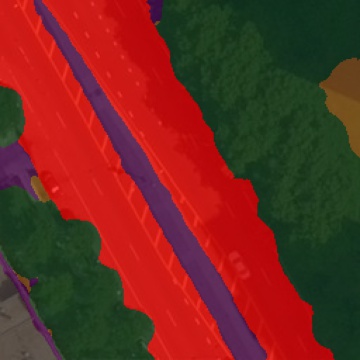} &
      \includegraphics[width=\w]{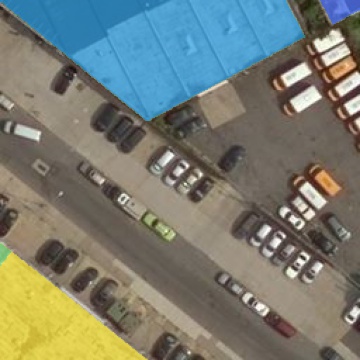} &
      \includegraphics[width=\w]{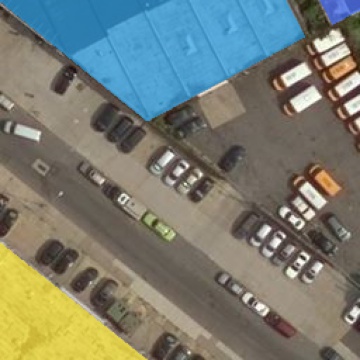} &
      \includegraphics[width=\w]{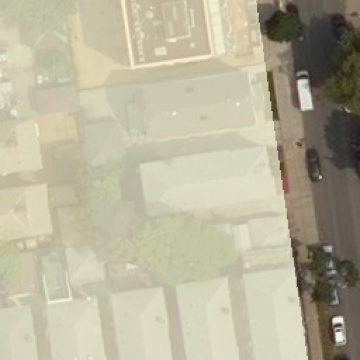} &
      \includegraphics[width=\w]{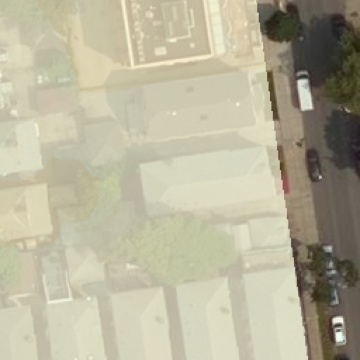} \\
      
      \includegraphics[width=\w]{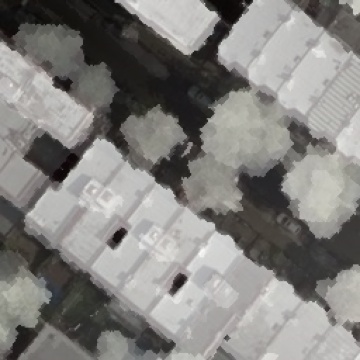} &
      \includegraphics[width=\w]{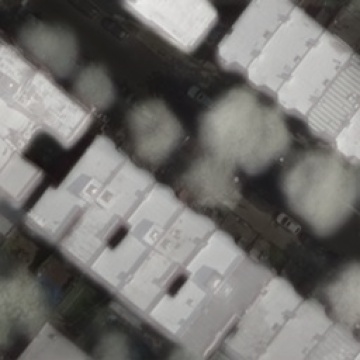} &
      \includegraphics[width=\w]{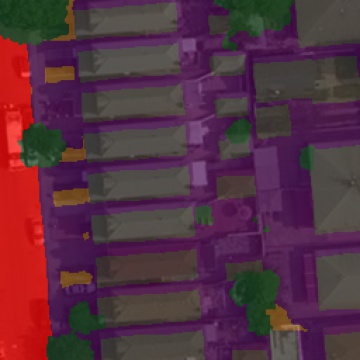} &
      \includegraphics[width=\w]{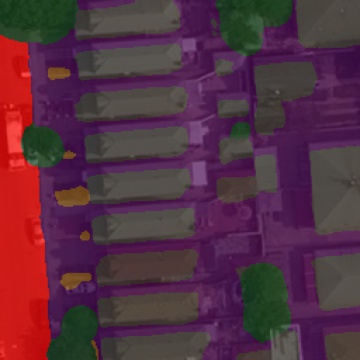} &
      \includegraphics[width=\w]{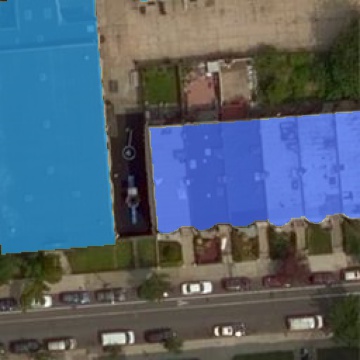} &
      \includegraphics[width=\w]{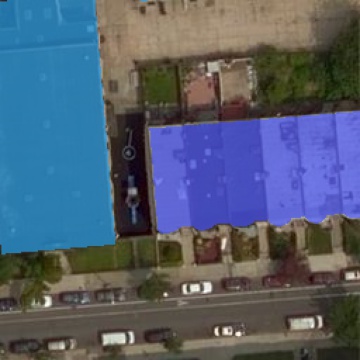} &
      \includegraphics[width=\w]{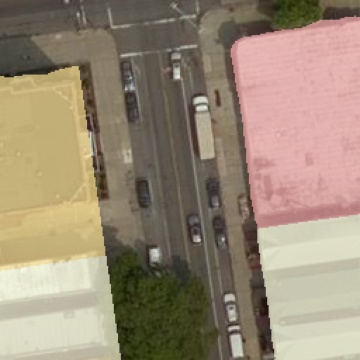} &
      \includegraphics[width=\w]{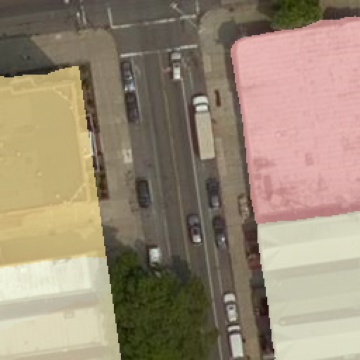} \\
      
      \includegraphics[width=\w]{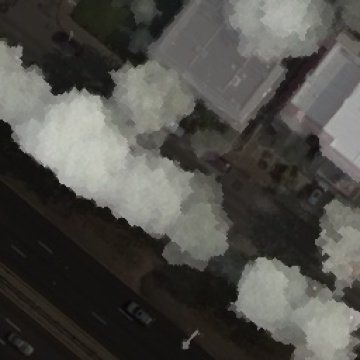} &
      \includegraphics[width=\w]{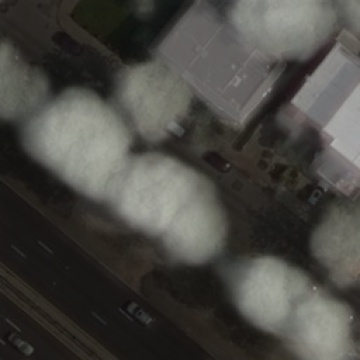} &
      \includegraphics[width=\w]{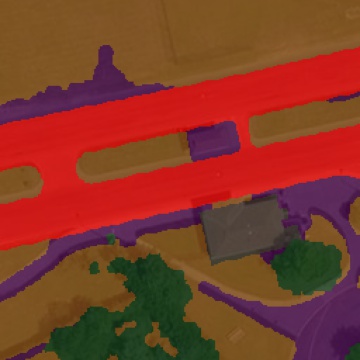} &
      \includegraphics[width=\w]{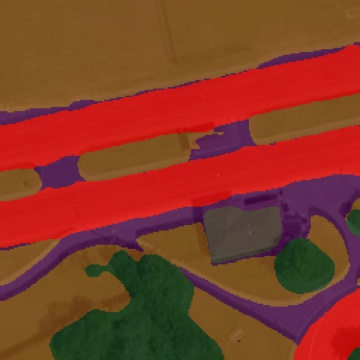} &
      \includegraphics[width=\w]{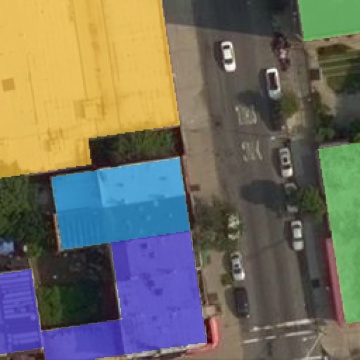} &
      \includegraphics[width=\w]{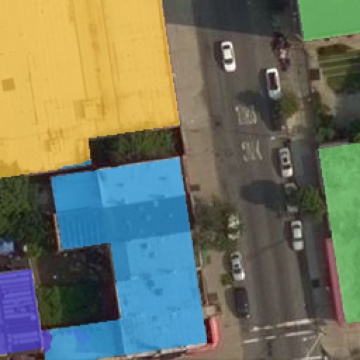} &
      \includegraphics[width=\w]{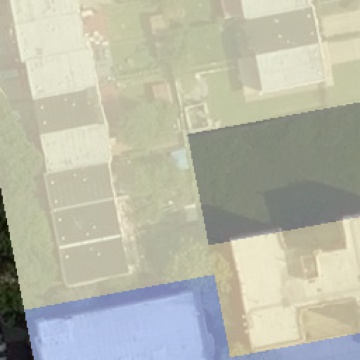} &
      \includegraphics[width=\w]{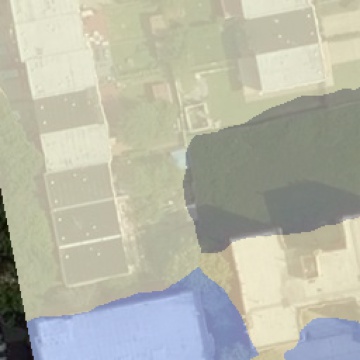} \\
      
      \includegraphics[width=\w]{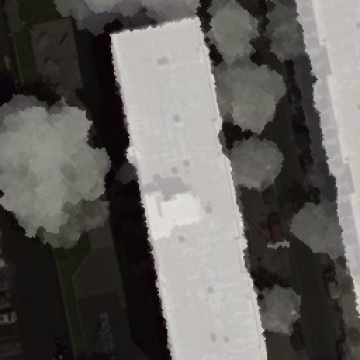} &
      \includegraphics[width=\w]{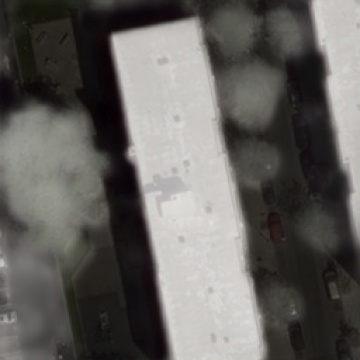} &
      \includegraphics[width=\w]{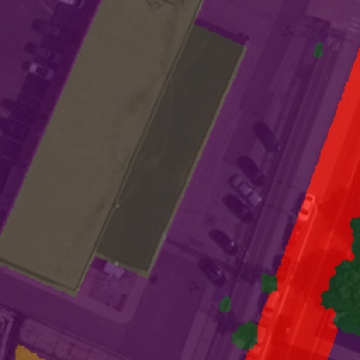} &
      \includegraphics[width=\w]{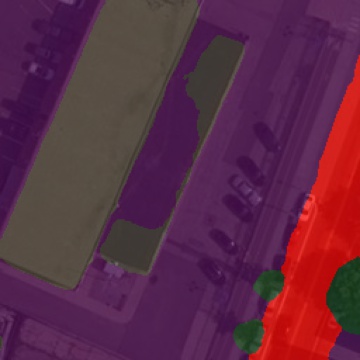} &
      \includegraphics[width=\w]{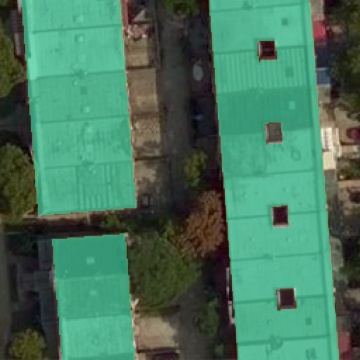} &
      \includegraphics[width=\w]{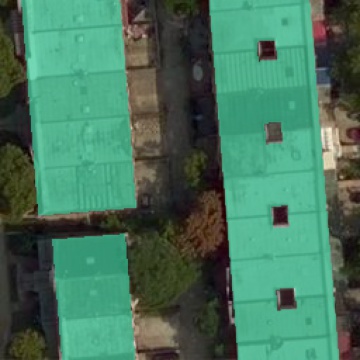} &
      \includegraphics[width=\w]{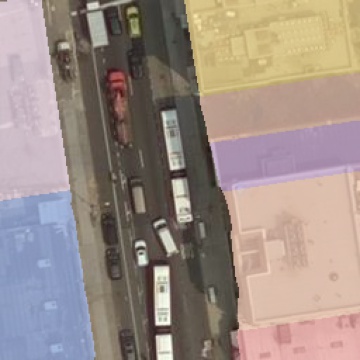} &
      \includegraphics[width=\w]{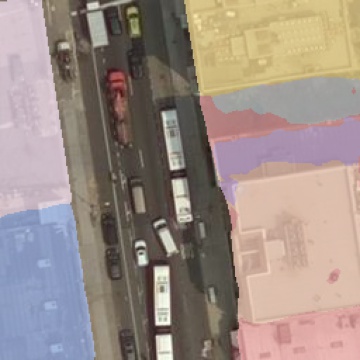} \\
      
      \includegraphics[width=\w]{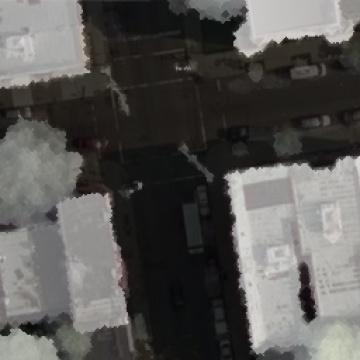} &
      \includegraphics[width=\w]{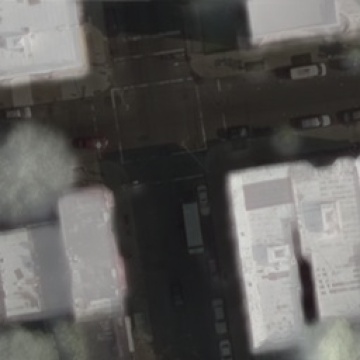} &
      \includegraphics[width=\w]{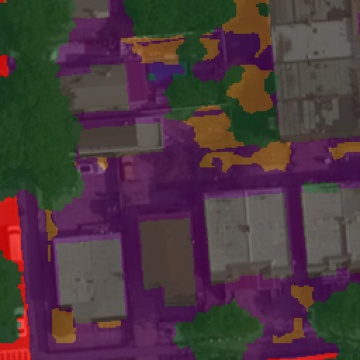} &
      \includegraphics[width=\w]{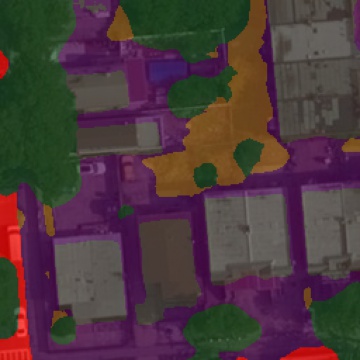} &
      \includegraphics[width=\w]{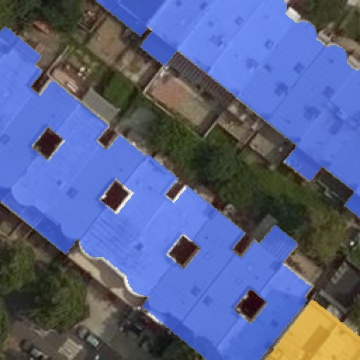} &
      \includegraphics[width=\w]{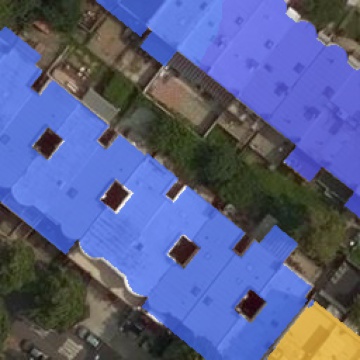} &
      \includegraphics[width=\w]{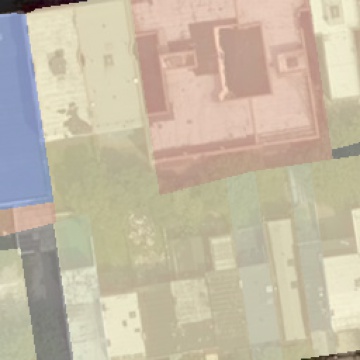} &
      \includegraphics[width=\w]{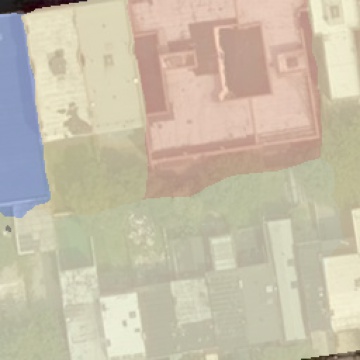} \\
      
      \includegraphics[width=\w]{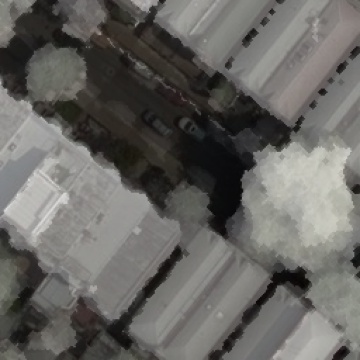} &
      \includegraphics[width=\w]{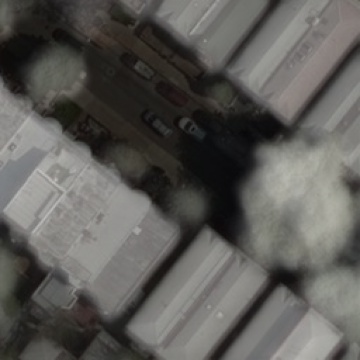} &
      \includegraphics[width=\w]{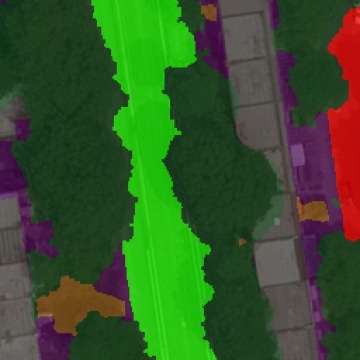} &
      \includegraphics[width=\w]{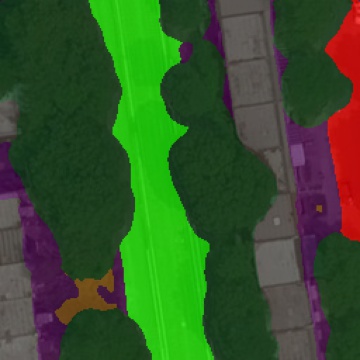} &
      \includegraphics[width=\w]{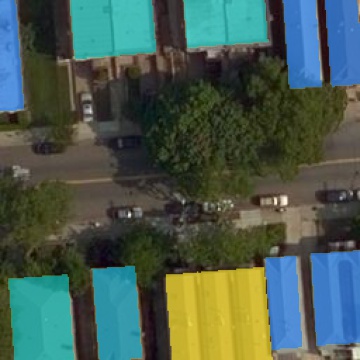} &
      \includegraphics[width=\w]{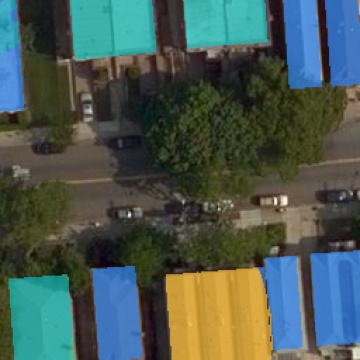} &
      \includegraphics[width=\w]{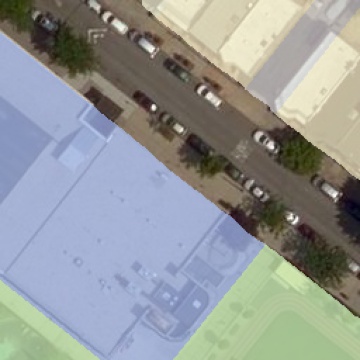} &
      \includegraphics[width=\w]{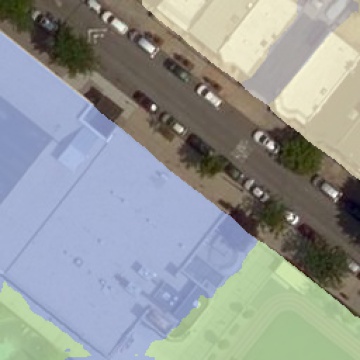} \\
      
    \end{tabular}

  \caption{Example qualitative results: (left) ground truth and (right) {\em ours}.}

  \label{fig:qualitative_results}

\end{figure*}

\tabref{brooklyn_results} shows quantitative results for our method on all five tasks versus baselines. Our full method ({\em ours}), significantly outperforms the single-modality baselines ({\em proximate} and {\em remote}) that are built from components of our method. Similarly, our approach, which integrates geospatial attention, outperforms two prior methods on the subset of tasks where results for the respective method were available. In addition, these results demonstrate that integrating nearby ground-level imagery using our method can even benefit tasks such as land cover estimation, where overhead imagery tends to be sufficient. \figref{qualitative_results} shows qualitative results.

\begin{table}
  \centering
  \caption{Performance analysis when upgrading components of a baseline~\cite{workman2017unified} to be similar to our approach (Brooklyn land use estimation). The inclusion of geospatial attention results in the largest performance gain.}
  
  \resizebox{\linewidth}{!}{
      \begin{tabular}{@{}cccccc@{}}
        \toprule
        Method & Seg Arch & Pano Arch & \multicolumn{1}{c}{mIOU} & \multicolumn{1}{c}{Acc} \\
        \bottomrule
        \cite{workman2017unified} & {\em PixelNet} & {\em VGG-16} & 45.54\% & 77.40\% \\
        \cite{workman2017unified} & {\em LinkNet34} & {\em VGG-16} & 47.14\% & 76.52\% \\
        \cite{workman2017unified} & {\em LinkNet34} & {\em ResNet-50} & 51.59\% & 78.68\% \\
        ours & {\em LinkNet34} & {\em ResNet-50} & 67.43\% & 86.41\% \\
        ours & {\em EfficientUNet} & {\em ResNet-50} & \textbf{69.24\%} & \textbf{86.82\%} \\
        \bottomrule
      \end{tabular}
  }
  \label{tbl:update}
\end{table}

\begin{table}
  \centering
  \caption{Ablation study highlighting the importance of different input features for geospatial attention (Brooklyn land use estimation).}
  \begin{tabular}{@{}cccrr@{}}
    \toprule
    Panorama & Overhead & Geometry & \multicolumn{1}{c}{mIOU} & \multicolumn{1}{c}{Acc} \\
    \bottomrule
    \checkmark & & & 53.77\% & 79.38\% \\
    & \checkmark & & 53.66\% & 79.50\% \\
    & & $d$ & 59.41\% & 82.05\% \\
    & & $\theta$ & 62.04\% & 83.14\% \\
    & & $d, \theta$ & 68.46\% & 86.52\% \\
    \checkmark & & $d, \theta$ & 68.72\% & 86.58\% \\
    & \checkmark & $d, \theta$ & 68.87\% & 86.74\% \\
    \hline
    \checkmark & \checkmark & $d, \theta$  & \textbf{69.24\%} & \textbf{86.82\%} \\
    \bottomrule
  \end{tabular}
  \label{tbl:brooklyn_ablation}
\end{table}

Next, we analyze how much performance can be attributed to the use of geospatial attention to form the grid. For this, we update a previous method~\cite{workman2017unified} to use our segmentation architecture and a similar ground-level feature extractor (ResNet-50, after the global average pooling layer), while retaining their strategy for estimating a dense grid of features (locally weighted averaging w/ an adaptive kernel). \tabref{update} shows the results of this experiment. Though these upgrades do increase performance, the largest gains are due to our grid formulation (bottom two rows), which uses geospatial attention to extract geo-informative features from the ground-level images.

Finally, we conduct an ablation study in \tabref{brooklyn_ablation} to highlight the importance of the different input features used for geospatial attention. For this experiment, we focus on the land use task and compare performance using different combinations of inputs (geometry, panorama, overhead) when estimating the spatial attention map. Note that the panorama-only variant is essentially traditional attention, comparable to CBAM~\cite{woo2018cbam}. Our full approach outperforms all baselines, with the geometric features being essential for achieving good performance.

\subsection{Visualizing Geospatial Attention}

\begin{figure*}
    \centering

    \begin{subfigure}{.177\linewidth}
        \centering
        
        \setlength\tabcolsep{1pt}
    
        \begin{tabular}{c}
          \\
          \includegraphics[width=1\linewidth]{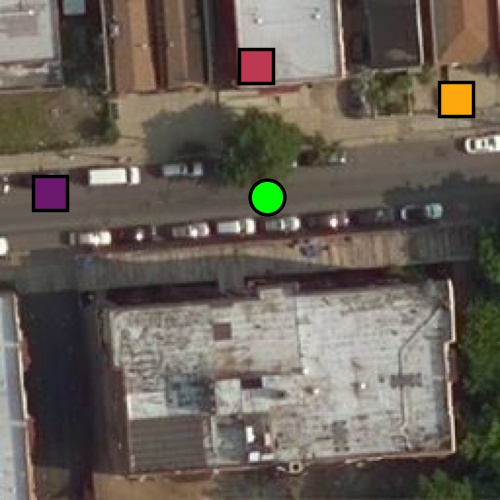} \\
          \includegraphics[width=1\linewidth]{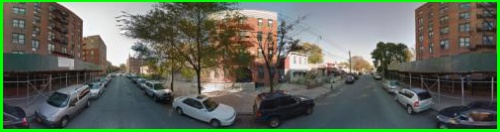} \\
        \end{tabular}
        
    \end{subfigure}
    \hfill
    \begin{subfigure}{.815\linewidth}
        \centering
        
        \setlength\tabcolsep{1pt}
    
        \begin{tabular}{ccc}
        
          Panorama & Geometry ($d, \theta$) & Ours \\
            
          \includegraphics[width=.32\linewidth]{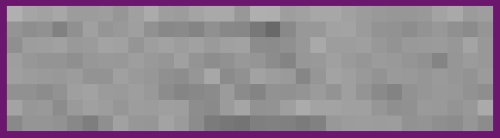} &
          \includegraphics[width=.32\linewidth]{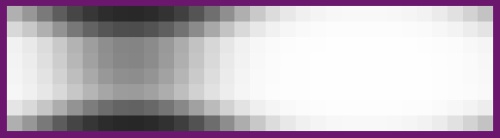} &
          \includegraphics[width=.32\linewidth]{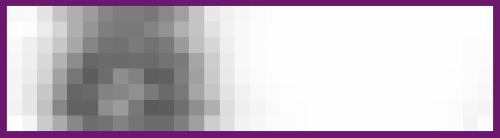} \\
          
          \includegraphics[width=.32\linewidth]{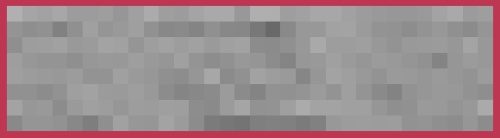} &
          \includegraphics[width=.32\linewidth]{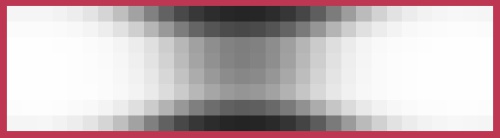} &
          \includegraphics[width=.32\linewidth]{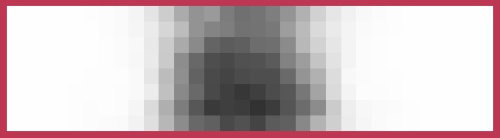} \\
          
          \includegraphics[width=.32\linewidth]{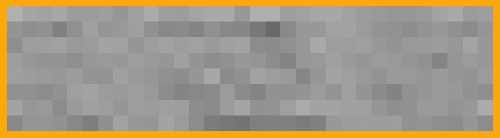} &
          \includegraphics[width=.32\linewidth]{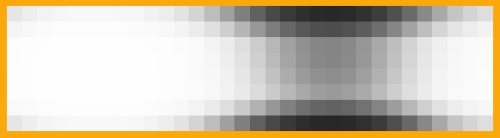} &
          \includegraphics[width=.32\linewidth]{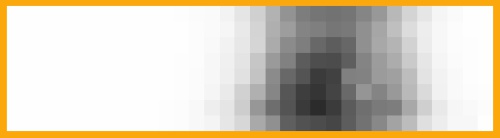} \\
        
        \end{tabular}

    \end{subfigure}

  \caption{Visualizing attention maps, learned as part of our land use ablation study, for a single panorama. The panorama location is represented by a green dot in the overhead image. Similarly, each row of attention maps is color-coded to correspond to one of three target pixels. The columns correspond respectively to the 1\textsuperscript{st}, 5\textsuperscript{th}, and bottom rows of \tabref{brooklyn_ablation}.}

  \label{fig:attention_types}

\end{figure*}

\begin{figure*}
    \centering

    \setlength\tabcolsep{1pt}

    \begin{tabular}{cccc}
    
      \includegraphics[height=.08\linewidth]{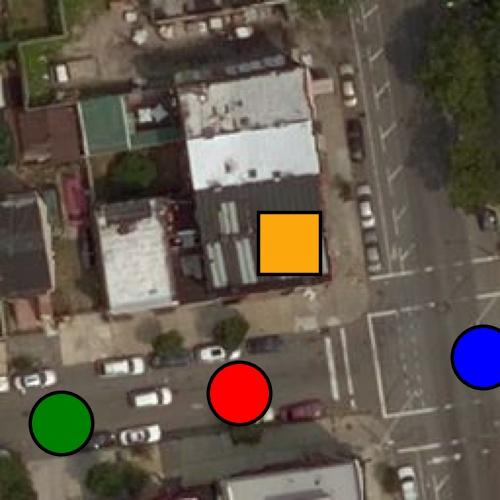} &
      \includegraphics[height=.08\linewidth]{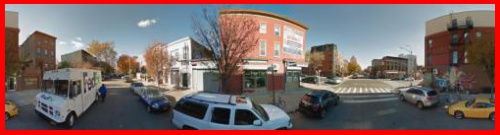} &
      \includegraphics[height=.08\linewidth]{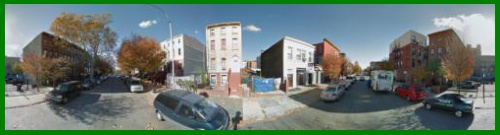} &
      \includegraphics[height=.08\linewidth]{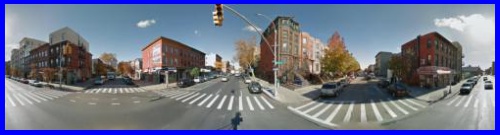} \\
      
      \includegraphics[height=.08\linewidth]{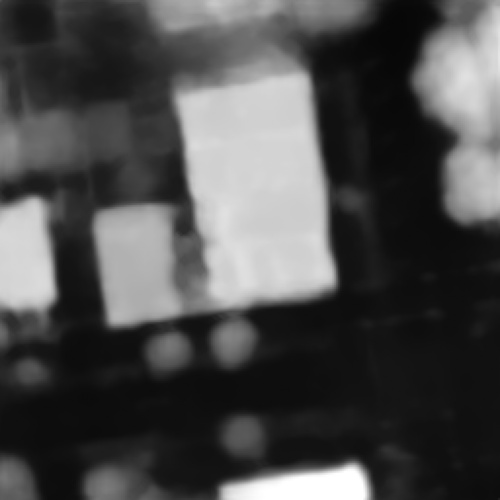} &
      \includegraphics[height=.08\linewidth]{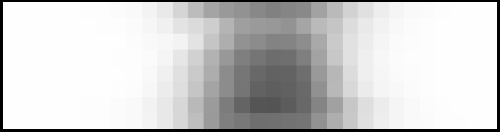} &
      \includegraphics[height=.08\linewidth]{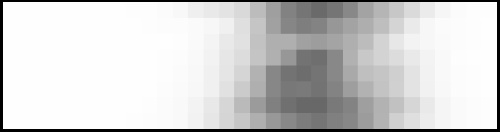} &
      \includegraphics[height=.08\linewidth]{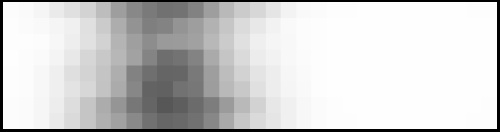} \\
      
      \includegraphics[height=.08\linewidth]{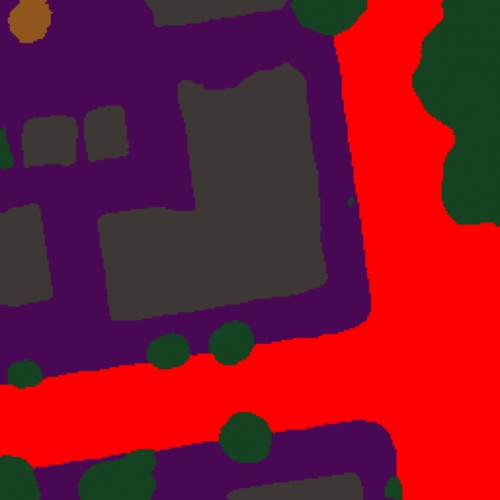} &
      \includegraphics[height=.08\linewidth]{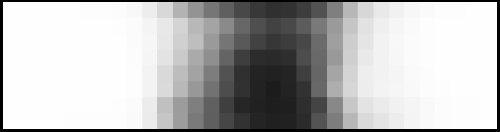} &
      \includegraphics[height=.08\linewidth]{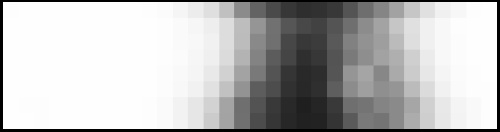} &
      \includegraphics[height=.08\linewidth]{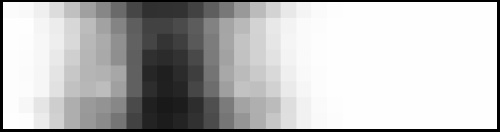} \\
      
      \includegraphics[height=.08\linewidth]{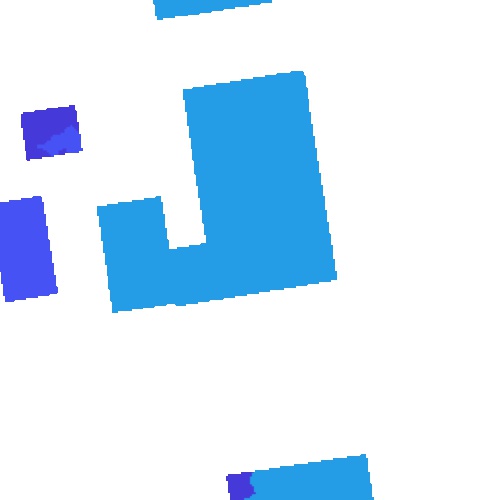} &
      \includegraphics[height=.08\linewidth]{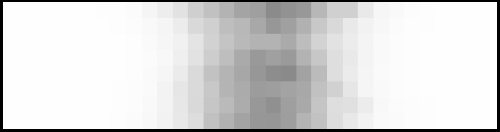} &
      \includegraphics[height=.08\linewidth]{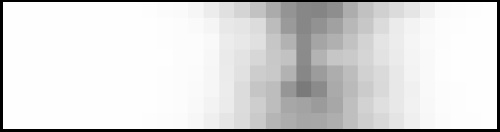} &
      \includegraphics[height=.08\linewidth]{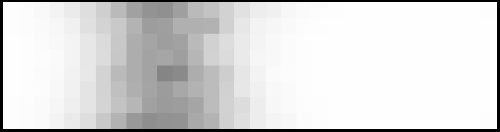} \\
    
    \end{tabular}

    \caption{Visualizing attention maps, for three panoramas (columns) and a single target pixel, learned for estimating height, land cover, and building age (rows). The panorama locations are shown in the overhead image as color-coded dots and the target pixel is shown as an orange square. While the attention maps focus on similar image regions, they are clearly task dependent.}

    \label{fig:attention_tasks}

\end{figure*}

\begin{figure}
    \centering
    
    \setlength{\fboxsep}{0pt}
    \setlength{\fboxrule}{1pt}
     
    \begin{subfigure}{.32\linewidth}
        \centering
        \fbox{\includegraphics[width=1\linewidth,trim=.6cm .6cm .6cm .6cm, clip]{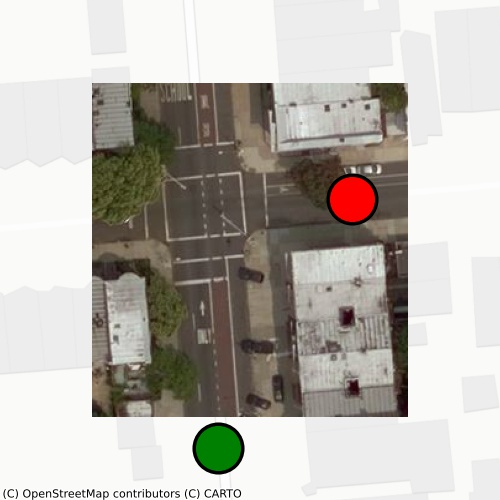}}
        \caption{Panorama Locations}
    \end{subfigure}
    \begin{subfigure}{.32\linewidth}
        \centering
        \fbox{\includegraphics[width=1\linewidth]{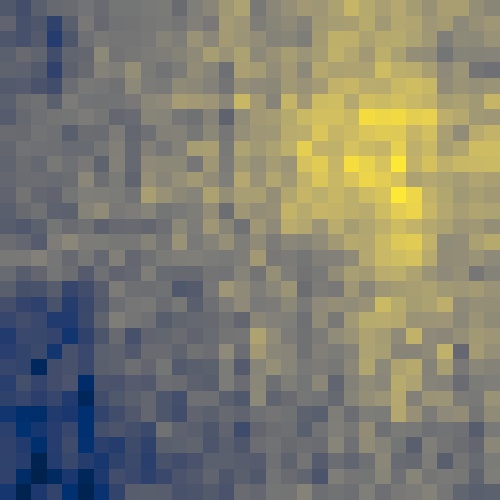}}
        \caption{Attention (red)}
    \end{subfigure}
    \begin{subfigure}{.32\linewidth}
        \centering
        \fbox{\includegraphics[width=1\linewidth]{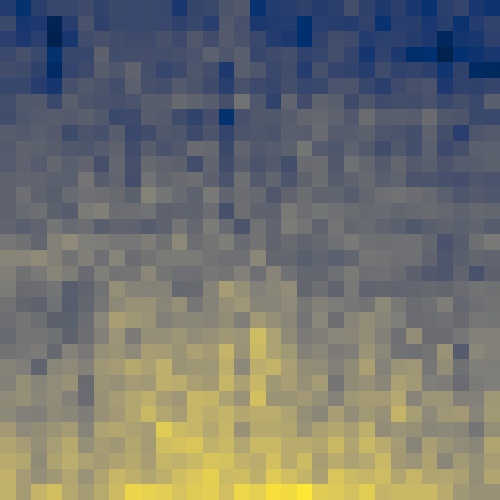}}
        \caption{Attention (green)}
    \end{subfigure}
    
    \caption{Visualizing the total attention in the overhead image for two panoramas (not shown but with locations shown as dots in the overhead image), where yellow (blue) means high (low) attention for the corresponding panorama. }

    \label{fig:attention_overhead}

\end{figure}

Geospatial attention is a flexible method for extracting information about a target location from a georegistered ground-level image. \figref{attention_types} shows qualitative examples of attention maps learned as part of our ablation study in \tabref{brooklyn_ablation}. Each row of attention maps is color-coded to correspond to one of three target pixels and the panorama location is represented by a green dot in the overhead image. The panorama-only attention maps are fairly uniform and not dependent on the target pixel location. The geometry-only attention maps are oriented toward the target pixel, but lack the refinement present in our full attention maps. For example, the top-right attention map assigns less weight to the uninformative pixels directly on the road.

\figref{attention_tasks} visualizes our full attention maps from several tasks (height, land cover, building age) for one target pixel, and three panoramas. As expected, they show that the region of high attention is generally oriented toward the target pixel. However, the region of the image that is attended depends on image content and attention changes depending on the task. Similarly, if the image and target locations are farther apart, the region of high activation shrinks, essentially narrowing the field of view.

\figref{attention_overhead} visualizes the total attention of two panoramas with respect to each location in the corresponding overhead image. As expected, each panorama contributes differently to each overhead image pixel, with generally more attention for pixels near the panorama location. 

\section{Conclusion}

We introduced the concept of geospatial attention, proposed an approach for estimating it, and used it to build an end-to-end architecture for near/remote sensing. Our approach enables joint inference between overhead imagery and nearby ground-level images in a manner that is ``geometry-aware''. To support evaluation, we extended an existing dataset to include two new per-pixel labeling tasks, estimating height and land cover. Extensive experiments, including an ablation study, demonstrate how integrating geospatial attention enables superior performance versus baselines on five different tasks. A key limitation of our method is that it requires georeferenced ground-level panoramas. Though it fails gracefully when such imagery isn’t present (reverting to an overhead-only model), it would be ideal if our method was capable of incorporating imagery with unknown orientations, such as from social media.

{\footnotesize
\bibliographystyle{ieee_fullname}
\bibliography{biblio}
}

\null
\vskip .375in
\twocolumn[{%
  \begin{center}
    \textbf{\Large Supplemental Material : \\ Revisiting Near/Remote Sensing with Geospatial Attention}
  \end{center}
  \vspace*{24pt}
}]
\setcounter{section}{0}
\setcounter{equation}{0}
\setcounter{figure}{0}
\setcounter{table}{0}
\makeatletter
\renewcommand{\theequation}{S\arabic{equation}}
\renewcommand{\thefigure}{S\arabic{figure}}
\renewcommand{\thetable}{S\arabic{table}}

This document contains additional details and experiments related to
our methods.

\section{Dataset Details}

We extend the Brooklyn and Queens dataset~\cite{workman2017unified} with two new per-pixel labeling tasks, estimating land cover and estimating height. The original dataset contains non-overlapping overhead images downloaded from Bing Maps (zoom level 19, approximately 30~cm per pixel) and street-level panoramas from Google Street View. The Brooklyn subset consists of 43,605 overhead images and 139,327 panoramas. The held-out Queens subset, used solely for evaluation, consists of 10,044 overhead images and 38,603 panoramas. Including our two new tasks, there are five tasks for this dataset: estimating land use, building age, building function, land cover, and height. For all experiments, we include the 20 closest street-level panoramas to each overhead image. For evaluation, we use the original train/test splits. 

\section{Qualitative Results}

\begin{figure}

  \centering
 
  \includegraphics[width=.24\linewidth]{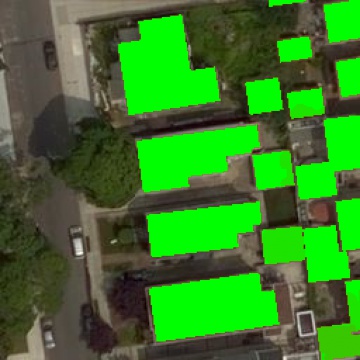}
  \includegraphics[width=.24\linewidth]{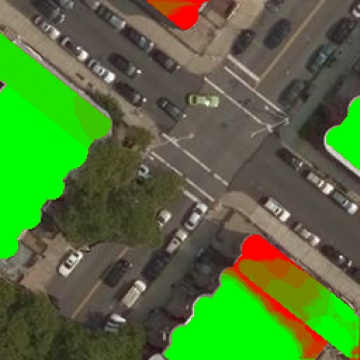}
  \includegraphics[width=.24\linewidth]{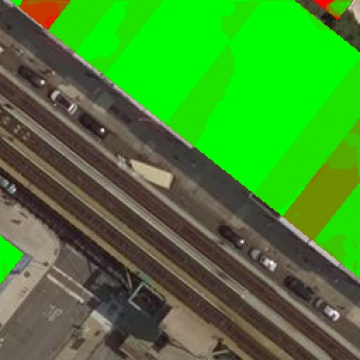}
  \includegraphics[width=.24\linewidth]{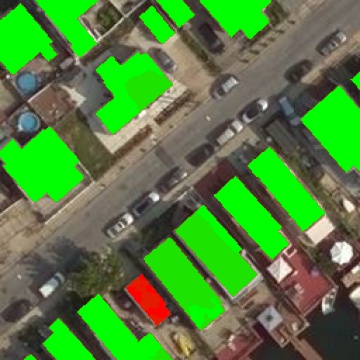}
  \includegraphics[width=.24\linewidth]{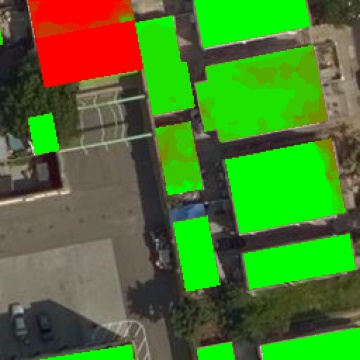}
  \includegraphics[width=.24\linewidth]{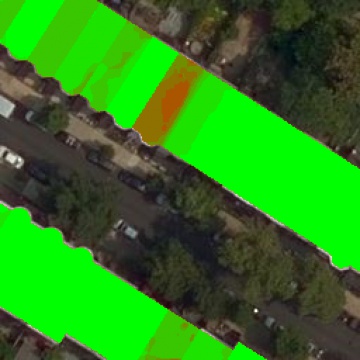}
  \includegraphics[width=.24\linewidth]{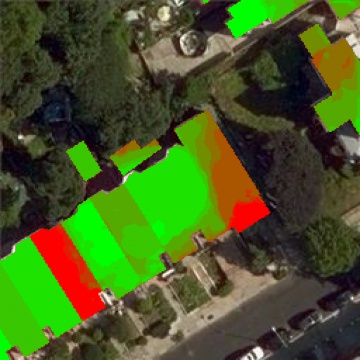}
  \includegraphics[width=.24\linewidth]{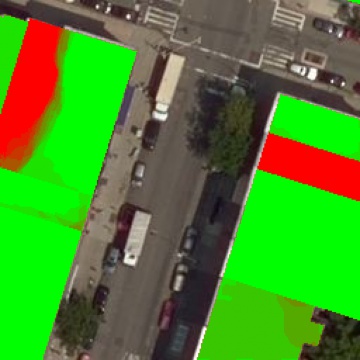}
 
  \caption{Qualitative results for building function. Each pixel represents the rank of the correct class in the posterior distribution (green to red). Bright green corresponds to rank one and red corresponds to rank 10 or more.}
  \label{fig:supp_qualitative_function}
\end{figure}

\begin{figure*}

  \centering
  
  \setlength\tabcolsep{1pt}
  \newcommand\w{.121\linewidth}    
    
    \begin{tabular}{cccccccc}
    
      \multicolumn{2}{c}{Height} & \multicolumn{2}{c}{Land Cover} & \multicolumn{2}{c}{Age} & \multicolumn{2}{c}{Land Use} \\
    
      \includegraphics[width=\w]{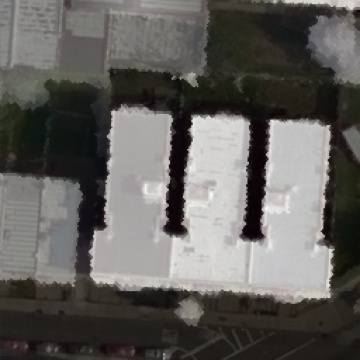} &
      \includegraphics[width=\w]{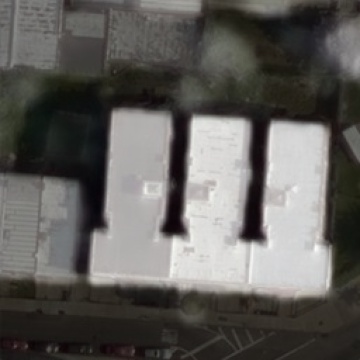} &
      \includegraphics[width=\w]{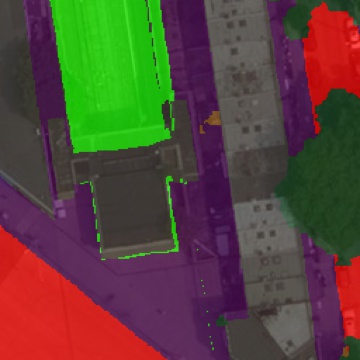} &
      \includegraphics[width=\w]{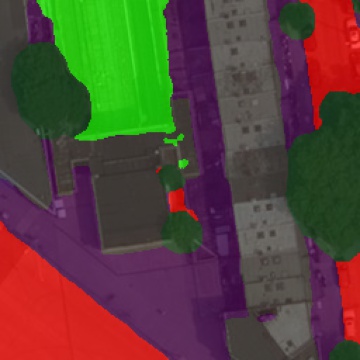} &
      \includegraphics[width=\w]{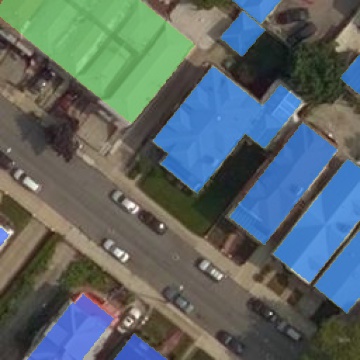} &
      \includegraphics[width=\w]{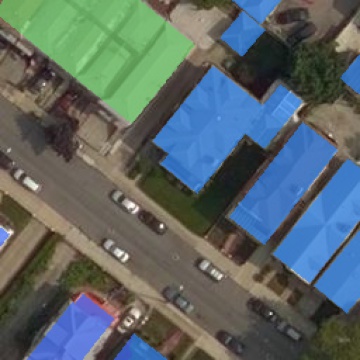} &
      \includegraphics[width=\w]{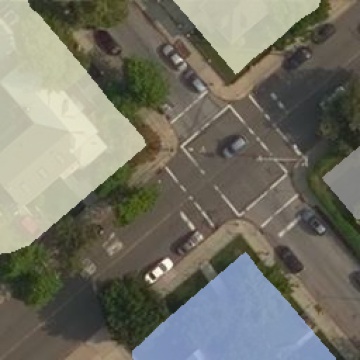} &
      \includegraphics[width=\w]{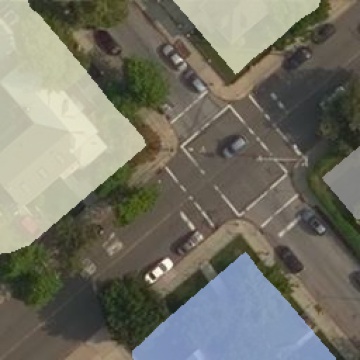} \\
      
      \includegraphics[width=\w]{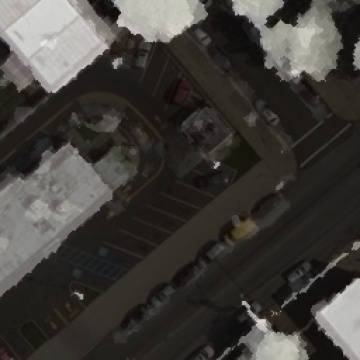} &
      \includegraphics[width=\w]{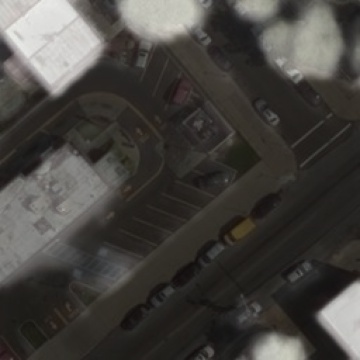} &
      \includegraphics[width=\w]{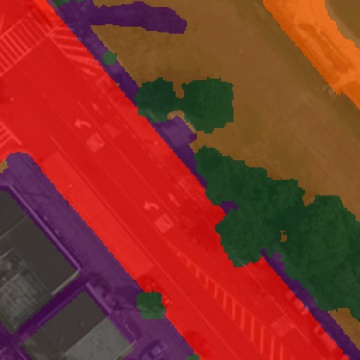} &
      \includegraphics[width=\w]{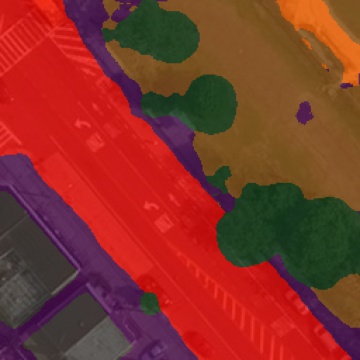} &
      \includegraphics[width=\w]{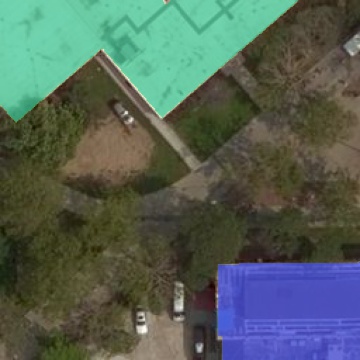} &
      \includegraphics[width=\w]{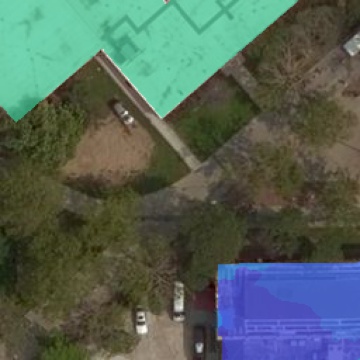} &
      \includegraphics[width=\w]{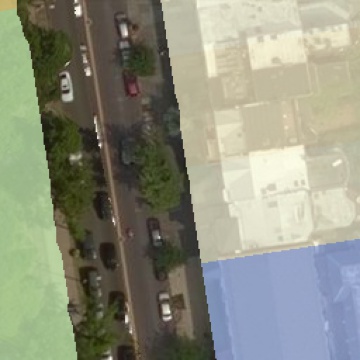} &
      \includegraphics[width=\w]{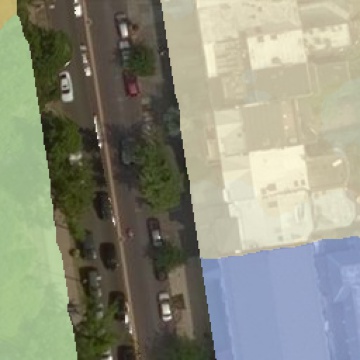} \\
      
      \includegraphics[width=\w]{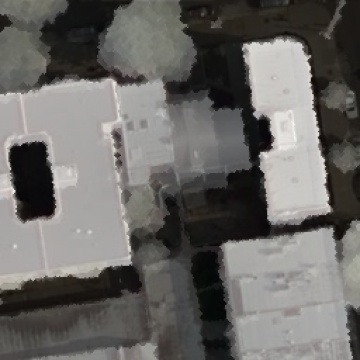} &
      \includegraphics[width=\w]{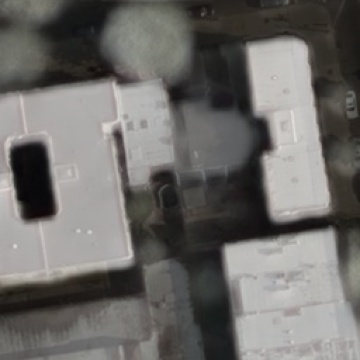} &
      \includegraphics[width=\w]{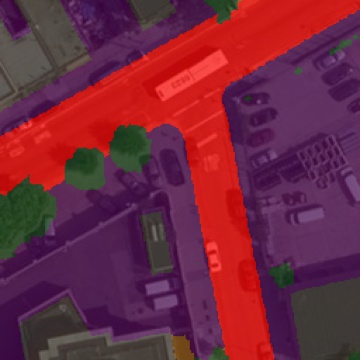} &
      \includegraphics[width=\w]{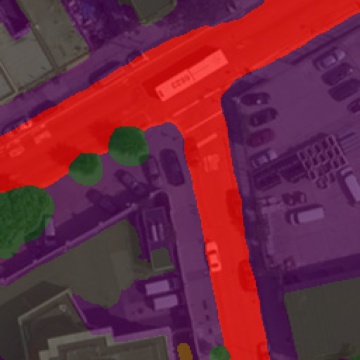} &
      \includegraphics[width=\w]{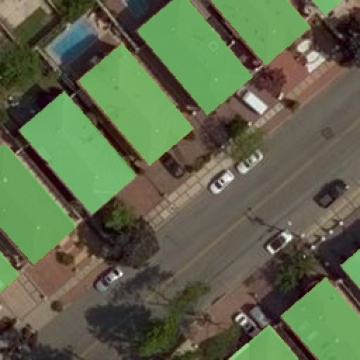} &
      \includegraphics[width=\w]{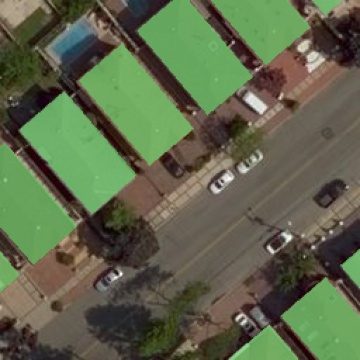} &
      \includegraphics[width=\w]{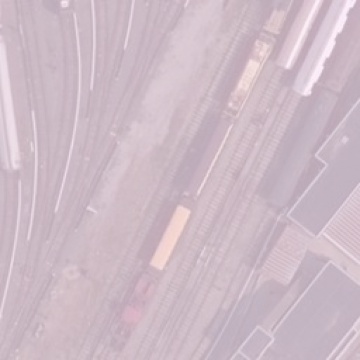} &
      \includegraphics[width=\w]{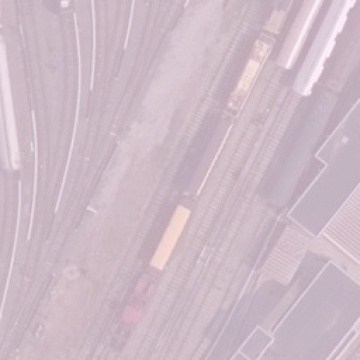} \\
      
      \includegraphics[width=\w]{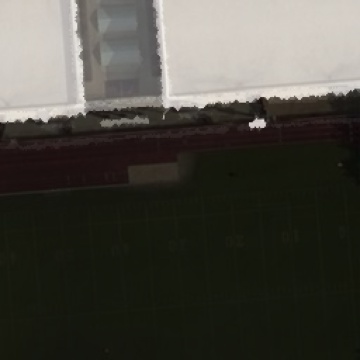} &
      \includegraphics[width=\w]{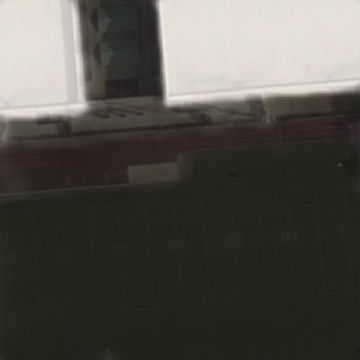} &
      \includegraphics[width=\w]{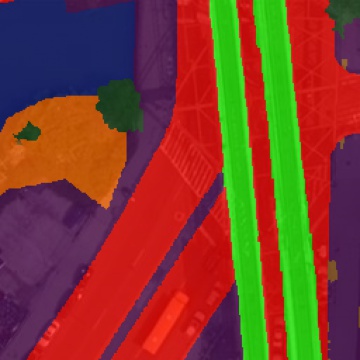} &
      \includegraphics[width=\w]{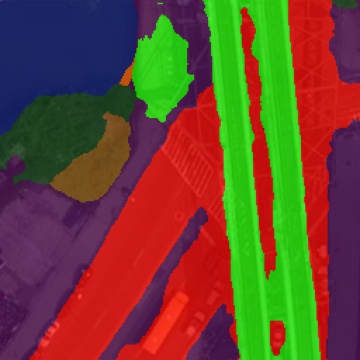} &
      \includegraphics[width=\w]{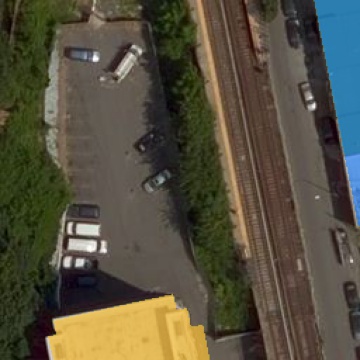} &
      \includegraphics[width=\w]{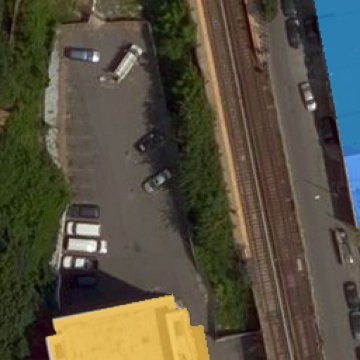} &
      \includegraphics[width=\w]{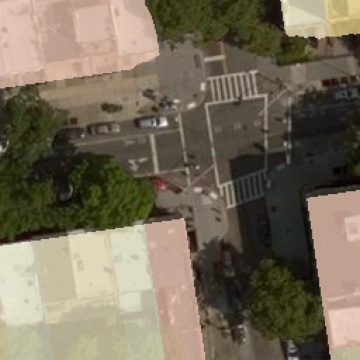} &
      \includegraphics[width=\w]{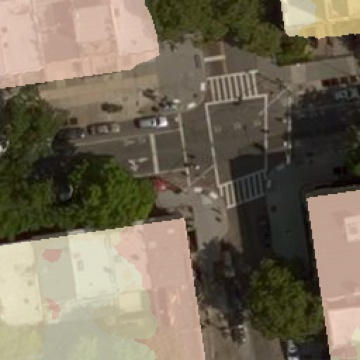} \\
      
      \includegraphics[width=\w]{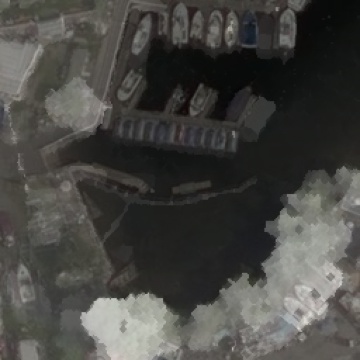} &
      \includegraphics[width=\w]{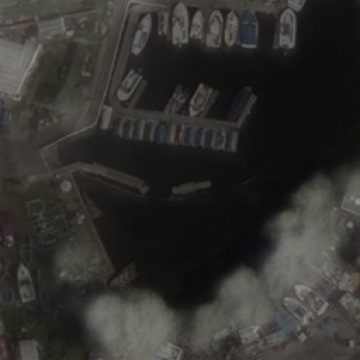} &
      \includegraphics[width=\w]{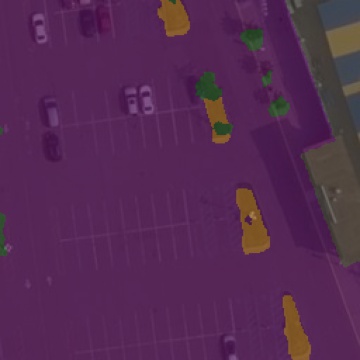} &
      \includegraphics[width=\w]{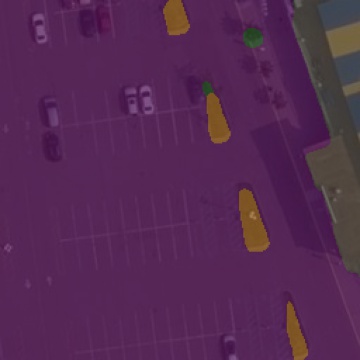} &
      \includegraphics[width=\w]{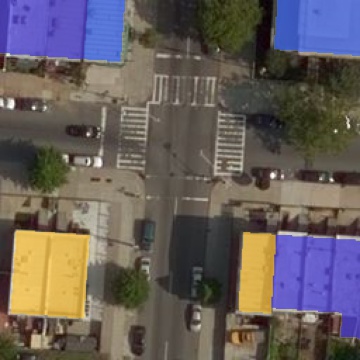} &
      \includegraphics[width=\w]{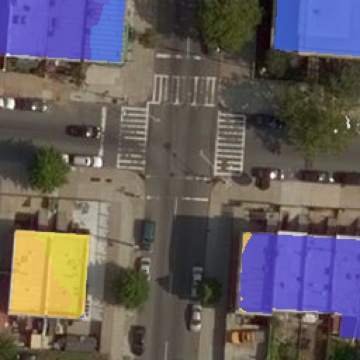} &
      \includegraphics[width=\w]{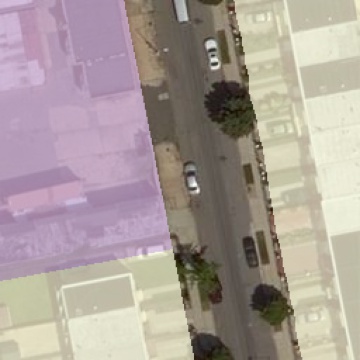} &
      \includegraphics[width=\w]{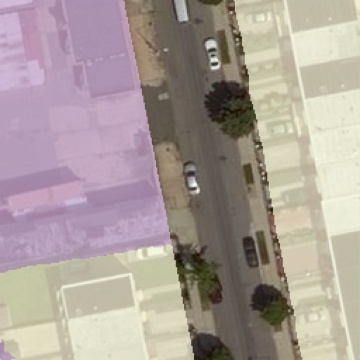} \\
      
      \includegraphics[width=\w]{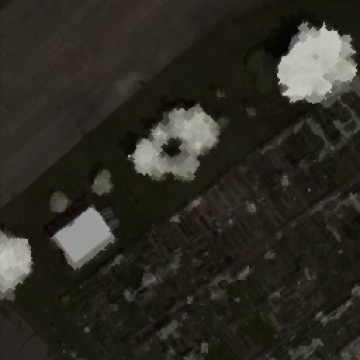} &
      \includegraphics[width=\w]{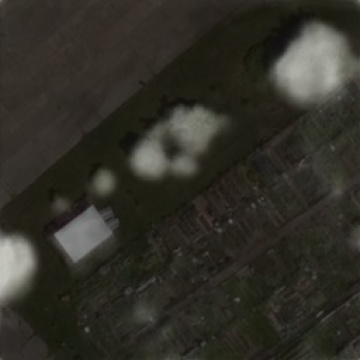} &
      \includegraphics[width=\w]{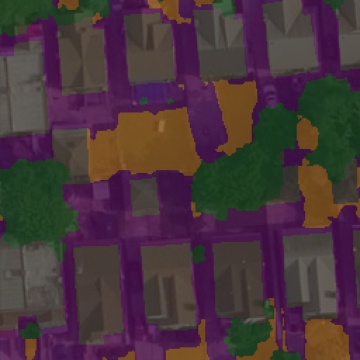} &
      \includegraphics[width=\w]{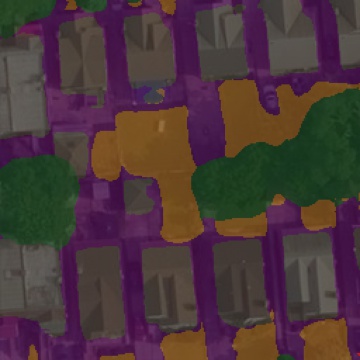} &
      \includegraphics[width=\w]{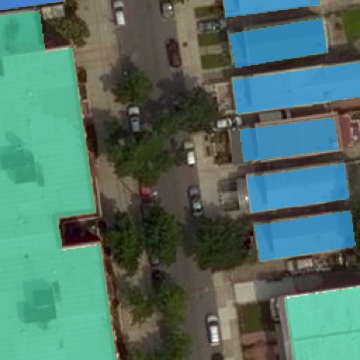} &
      \includegraphics[width=\w]{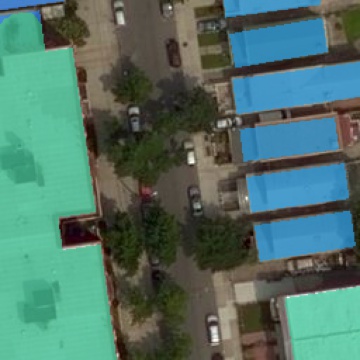} &
      \includegraphics[width=\w]{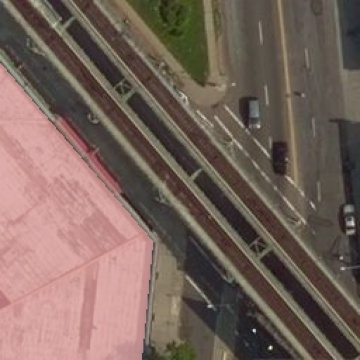} &
      \includegraphics[width=\w]{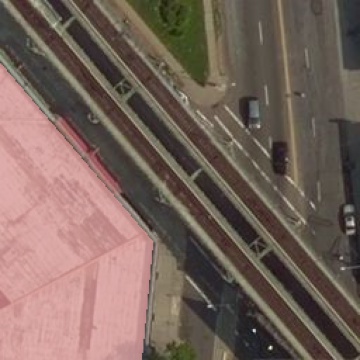} \\
      
      \includegraphics[width=\w]{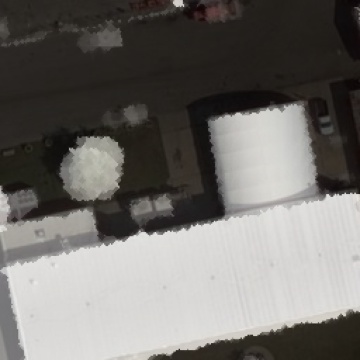} &
      \includegraphics[width=\w]{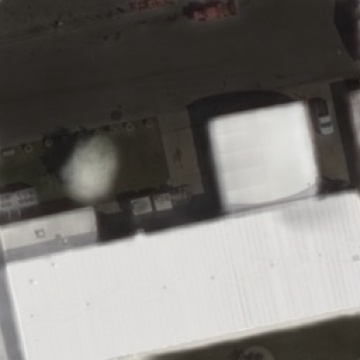} &
      \includegraphics[width=\w]{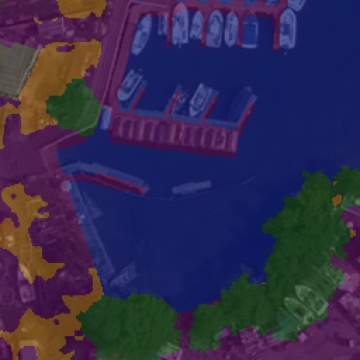} &
      \includegraphics[width=\w]{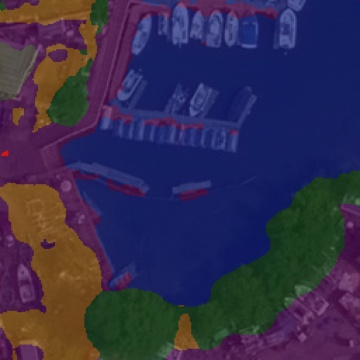} &
      \includegraphics[width=\w]{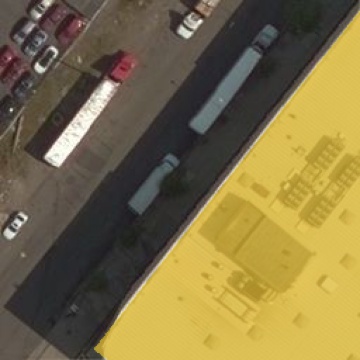} &
      \includegraphics[width=\w]{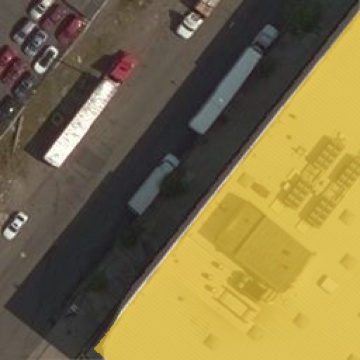} &
      \includegraphics[width=\w]{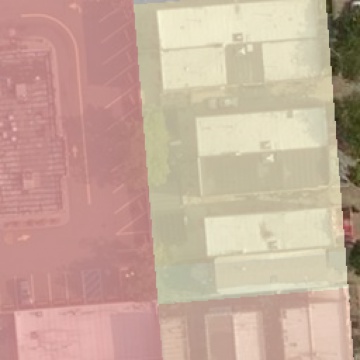} &
      \includegraphics[width=\w]{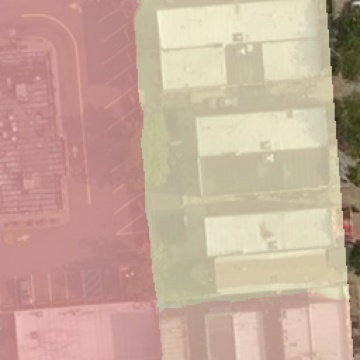} \\
      
      \includegraphics[width=\w]{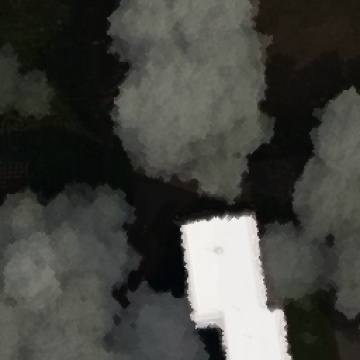} &
      \includegraphics[width=\w]{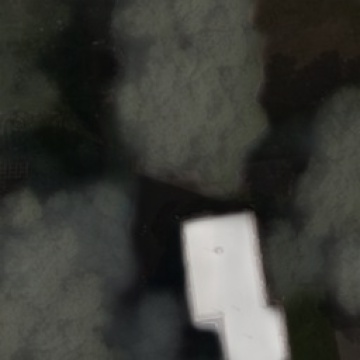} &
      \includegraphics[width=\w]{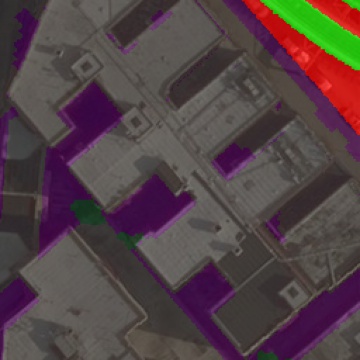} &
      \includegraphics[width=\w]{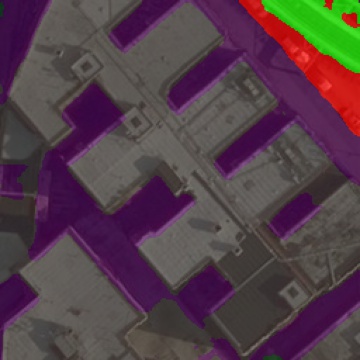} &
      \includegraphics[width=\w]{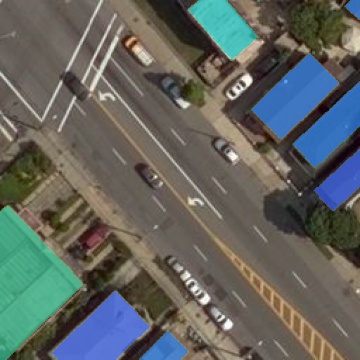} &
      \includegraphics[width=\w]{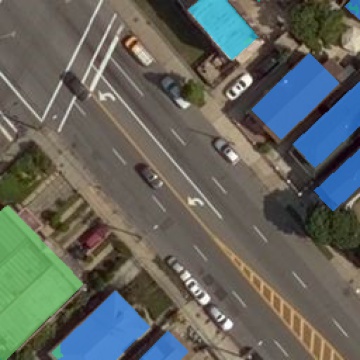} &
      \includegraphics[width=\w]{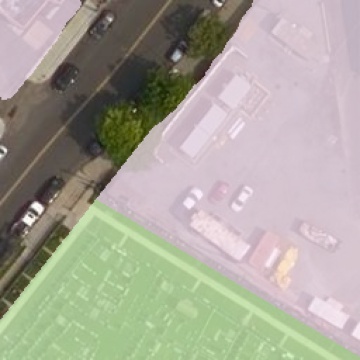} &
      \includegraphics[width=\w]{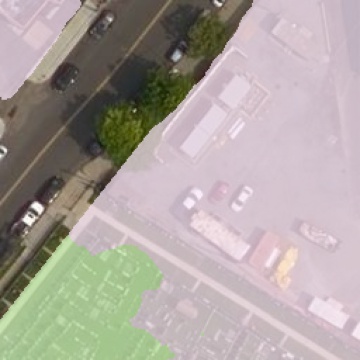} \\
      
      \includegraphics[width=\w]{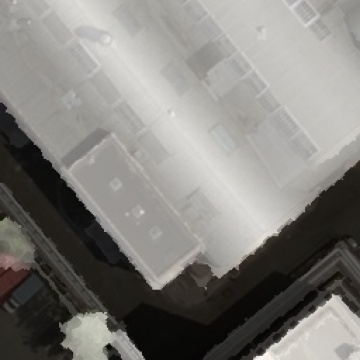} &
      \includegraphics[width=\w]{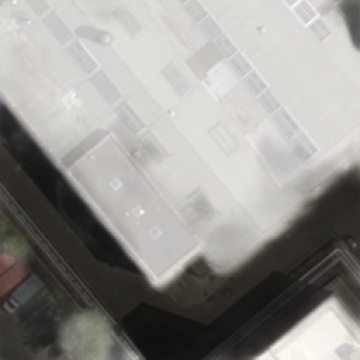} &
      \includegraphics[width=\w]{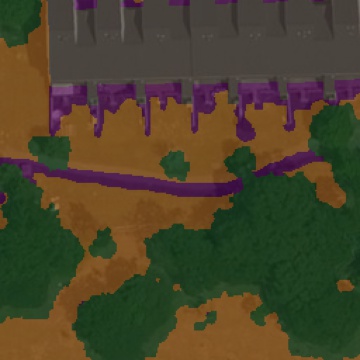} &
      \includegraphics[width=\w]{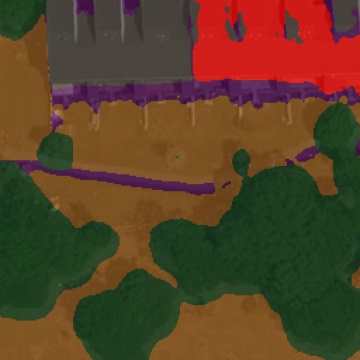} &
      \includegraphics[width=\w]{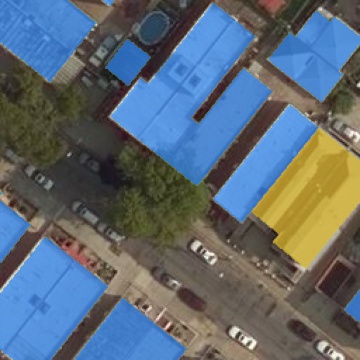} &
      \includegraphics[width=\w]{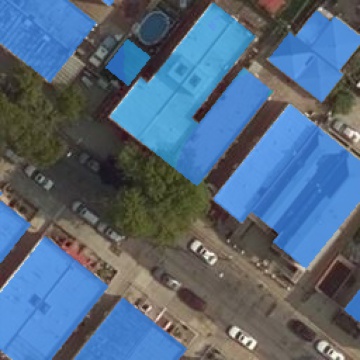} &
      \includegraphics[width=\w]{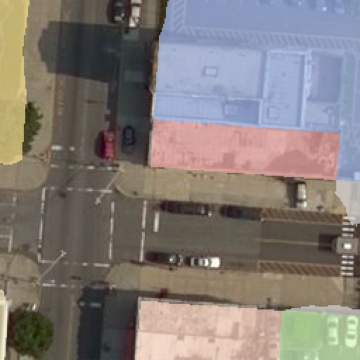} &
      \includegraphics[width=\w]{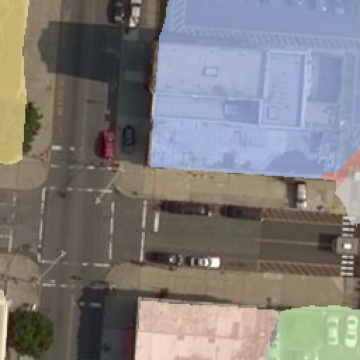} \\
    
    \end{tabular}

  \caption{Additional qualitative results: (left) ground truth and (right) {\em ours}.}

  \label{fig:supp_qualitative_results}

\end{figure*}

We show qualitative results for building function estimation in \figref{supp_qualitative_function}. Due the the large number of classes (206 building types), we visualize results for this task as a top-k image where each pixel is assigned a color (from green to red) by the rank of the correct class in the posterior distribution. Bright green corresponds to rank one and red corresponds rank 10 or more. We show additional qualitative results for the other tasks in \figref{supp_qualitative_results}.

\section{Attention Visualization}

\begin{figure*}
    \centering
    
    \begin{subfigure}{.32\linewidth}
        \centering
        \includegraphics[width=.49\linewidth]{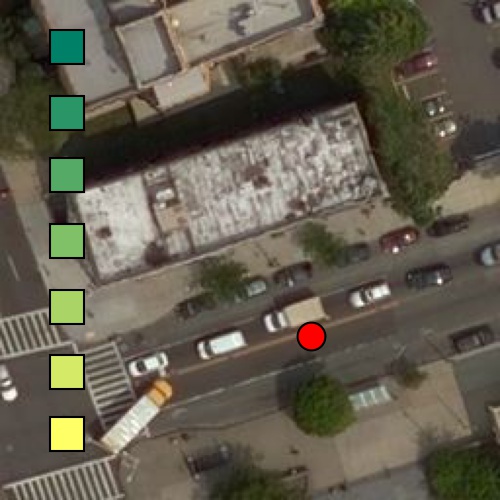}
        \includegraphics[width=.49\linewidth]{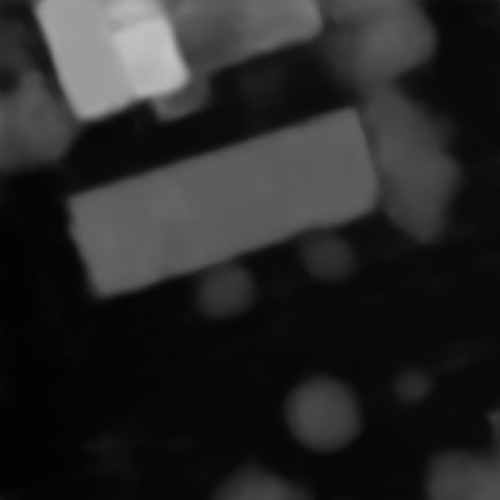}
        \includegraphics[width=1\linewidth]{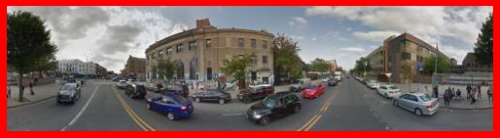}
    \end{subfigure}
    \hfill
    \begin{subfigure}{.32\linewidth}
        \centering
        \includegraphics[width=.49\linewidth]{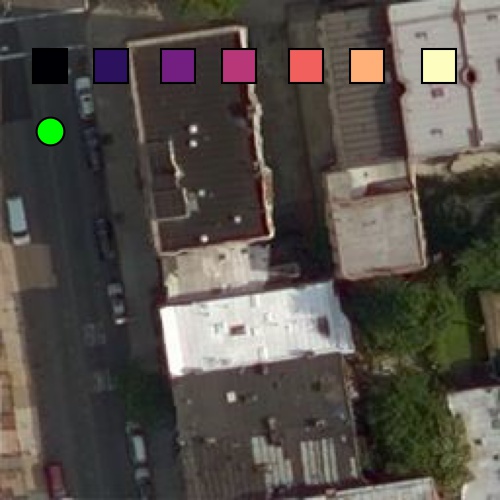}
        \includegraphics[width=.49\linewidth]{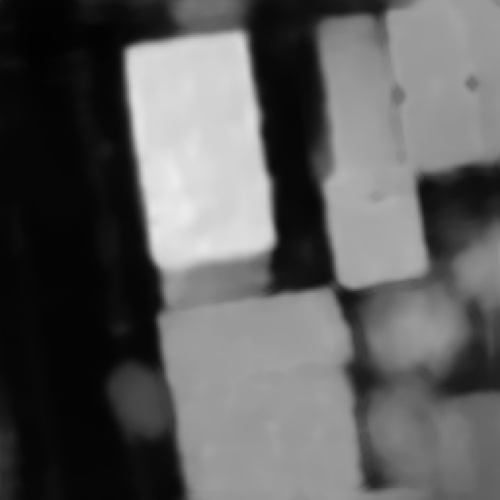}
        \includegraphics[width=1\linewidth]{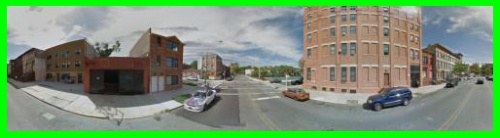}
    \end{subfigure}
    \hfill
    \begin{subfigure}{.32\linewidth}
        \centering
        \includegraphics[width=.49\linewidth]{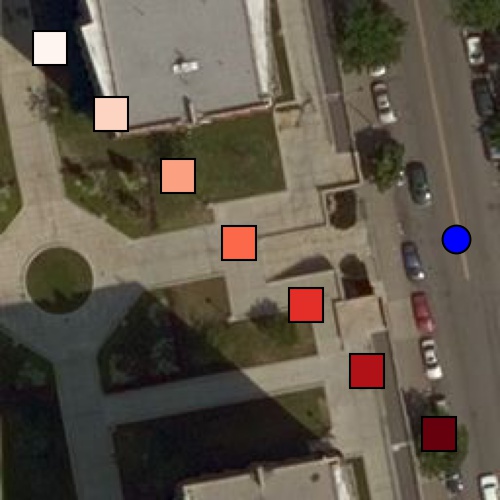}
        \includegraphics[width=.49\linewidth]{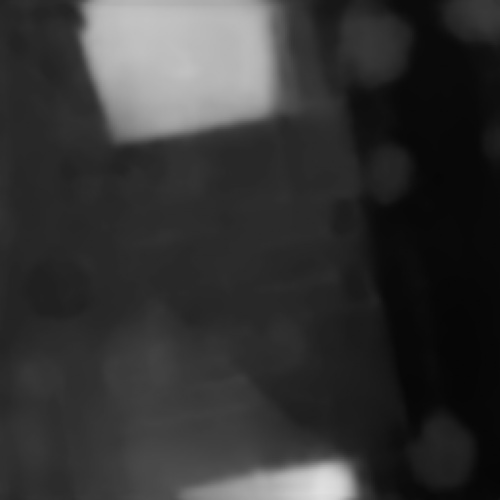}
        \includegraphics[width=1\linewidth]{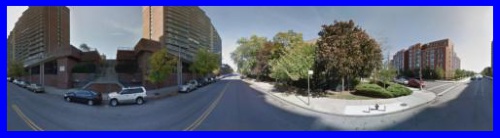}
    \end{subfigure}
    
    \par\bigskip
    \par\bigskip
    
    \begin{subfigure}{.32\linewidth}
        \centering
        \includegraphics[width=1\linewidth]{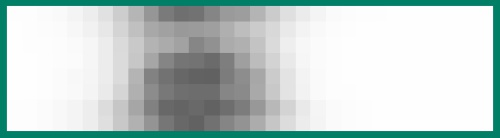}
        \includegraphics[width=1\linewidth]{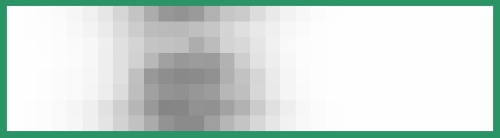}
        \includegraphics[width=1\linewidth]{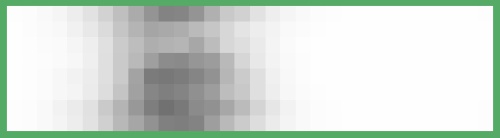}
        \includegraphics[width=1\linewidth]{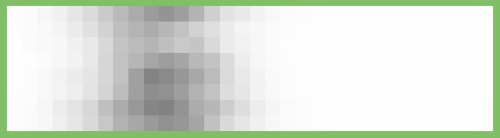}
        \includegraphics[width=1\linewidth]{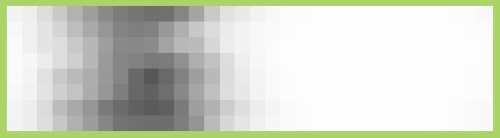}
        \includegraphics[width=1\linewidth]{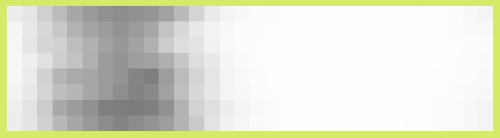}
        \includegraphics[width=1\linewidth]{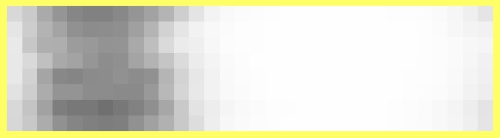}
    \end{subfigure}
    \hfill
    \begin{subfigure}{.32\linewidth}
        \centering
        \includegraphics[width=1\linewidth]{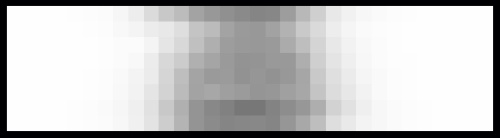}
        \includegraphics[width=1\linewidth]{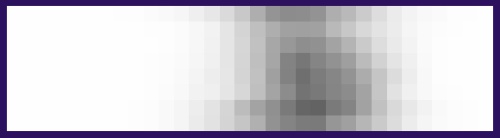}
        \includegraphics[width=1\linewidth]{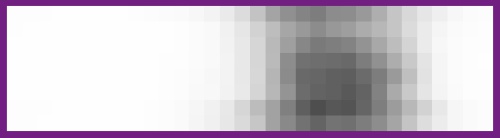}
        \includegraphics[width=1\linewidth]{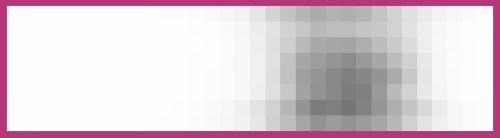}
        \includegraphics[width=1\linewidth]{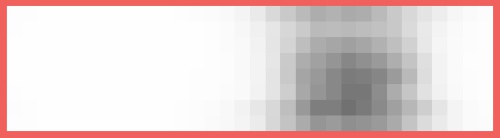}
        \includegraphics[width=1\linewidth]{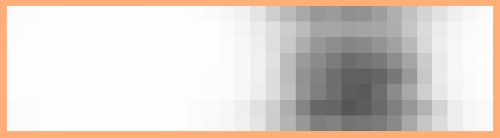}
        \includegraphics[width=1\linewidth]{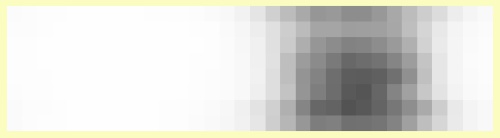}
    \end{subfigure}
    \hfill
    \begin{subfigure}{.32\linewidth}
        \centering
        \includegraphics[width=1\linewidth]{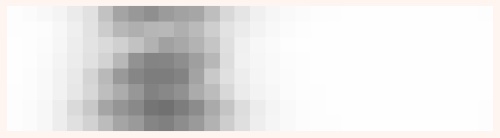}
        \includegraphics[width=1\linewidth]{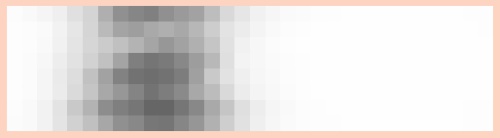}
        \includegraphics[width=1\linewidth]{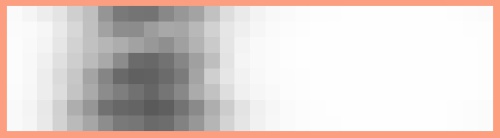}
        \includegraphics[width=1\linewidth]{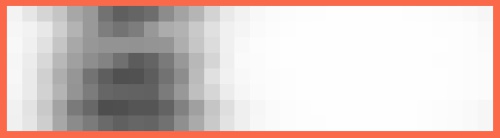}
        \includegraphics[width=1\linewidth]{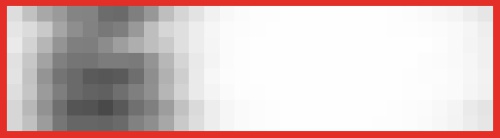}
        \includegraphics[width=1\linewidth]{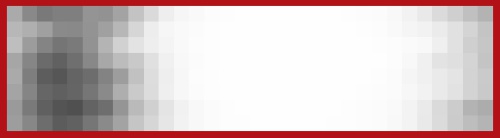}
        \includegraphics[width=1\linewidth]{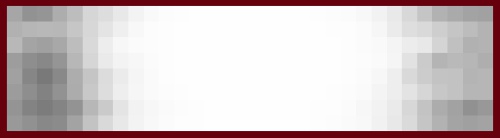}
    \end{subfigure}
    
  \caption{Visualizing spatial attention maps from our full method as the target location changes (height prediction task). Each column shows attention maps for one panorama, with the location of the panorama represented by the same-colored dot in the overhead image. Similarly, the attention maps are color-coded corresponding to the target location, which is represented by the same-colored square in the overhead image.}

  \label{fig:supp_attention_animation}

\end{figure*}

\figref{supp_attention_animation} visualizes the spatial attention maps for several input images as the target location changes. For this experiment, we use our full method and output from the height estimation task. Each image is color-coded and the capture location is represented by the same-colored dot in the overhead image. Similarly, the attention maps are color-coded, with the target location represented by the same-colored square in the overhead image. As observed, the region of high attention is generally oriented toward the target pixel. Our approach is able to learn these geometric relationships without requiring direct correspondences. 

\begin{figure*}
    \centering

    \begin{subfigure}{.1135\linewidth}
        \includegraphics[width=1\linewidth]{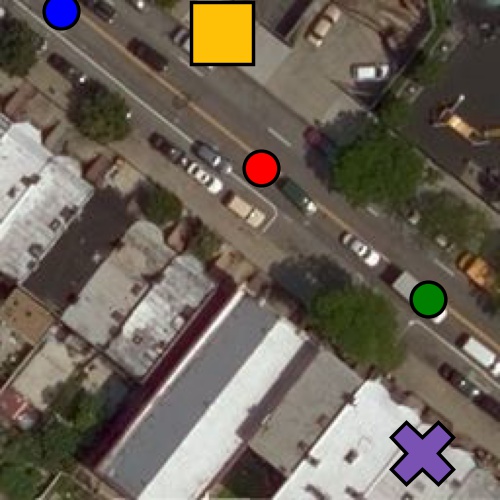}
        \includegraphics[width=1\linewidth]{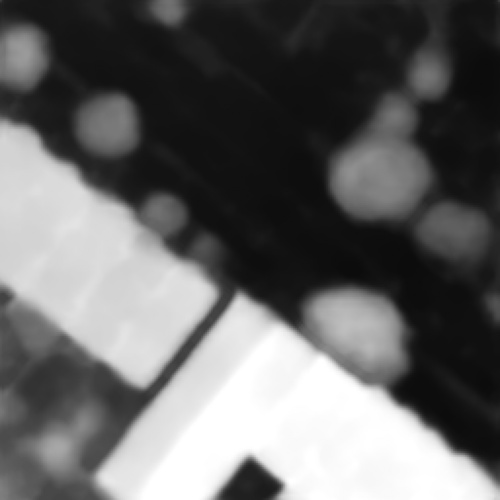}
    \end{subfigure}
    \begin{subfigure}{.29\linewidth}
        \includegraphics[width=1\linewidth]{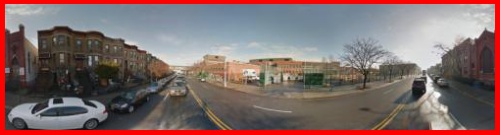}
        \includegraphics[width=1\linewidth]{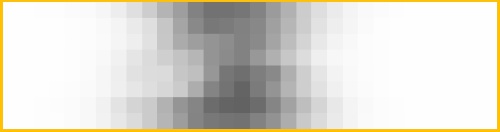}
        \includegraphics[width=1\linewidth]{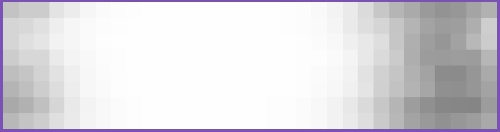}
    \end{subfigure}
    \begin{subfigure}{.29\linewidth}
        \includegraphics[width=1\linewidth]{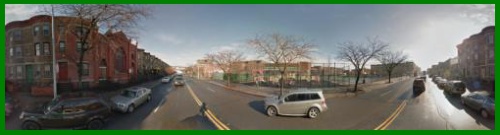}
        \includegraphics[width=1\linewidth]{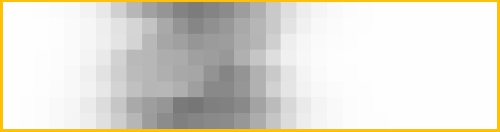}
        \includegraphics[width=1\linewidth]{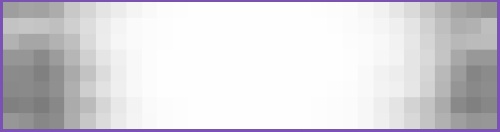}
    \end{subfigure}
    \begin{subfigure}{.29\linewidth}
        \includegraphics[width=1\linewidth]{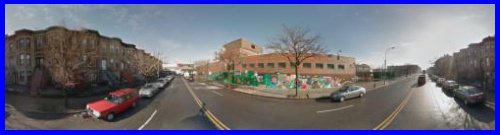}
        \includegraphics[width=1\linewidth]{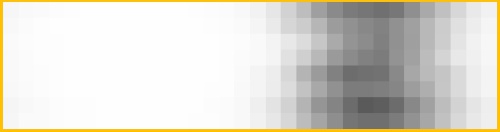}
        \includegraphics[width=1\linewidth]{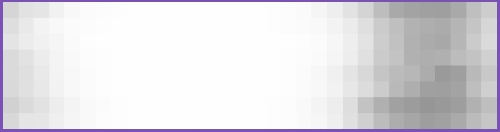}
    \end{subfigure}
    
    \par\medskip
    
    \begin{subfigure}{.1135\linewidth}
        \includegraphics[width=1\linewidth]{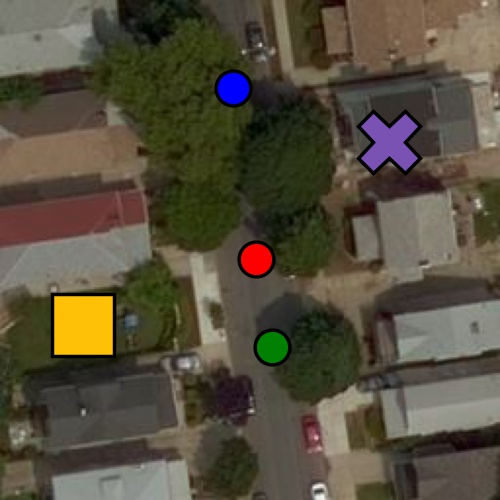}
        \includegraphics[width=1\linewidth]{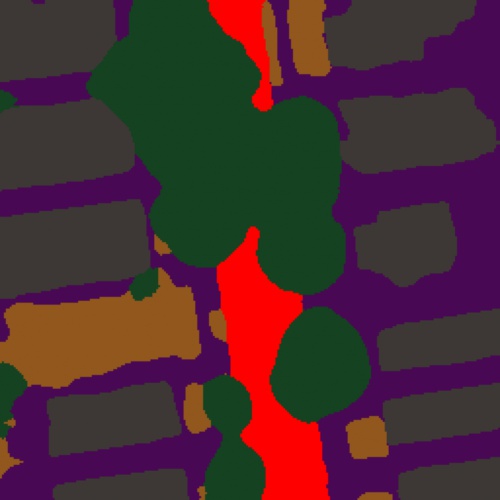}
    \end{subfigure}
    \begin{subfigure}{.29\linewidth}
        \includegraphics[width=1\linewidth]{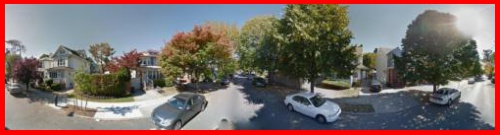}
        \includegraphics[width=1\linewidth]{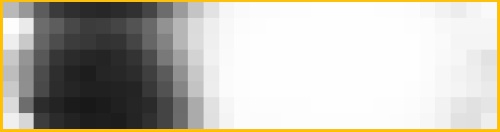}
        \includegraphics[width=1\linewidth]{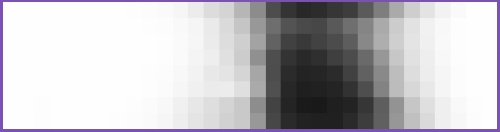}
    \end{subfigure}
    \begin{subfigure}{.29\linewidth}
        \includegraphics[width=1\linewidth]{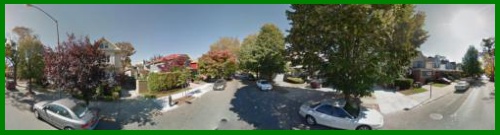}
        \includegraphics[width=1\linewidth]{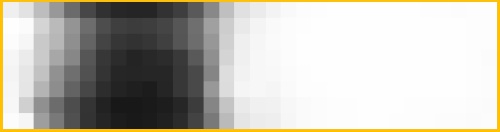}
        \includegraphics[width=1\linewidth]{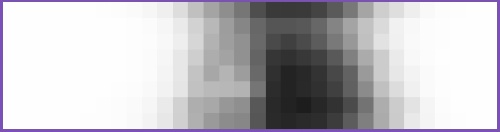}
    \end{subfigure}
    \begin{subfigure}{.29\linewidth}
        \includegraphics[width=1\linewidth]{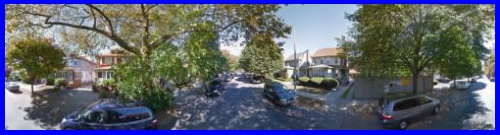}
        \includegraphics[width=1\linewidth]{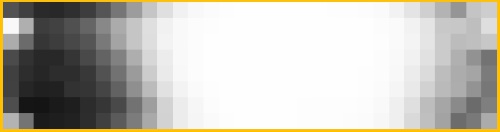}
        \includegraphics[width=1\linewidth]{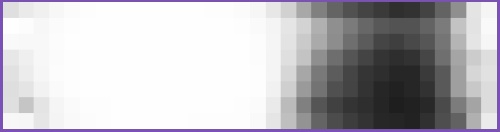}
    \end{subfigure}
    
    \par\medskip
    
    \begin{subfigure}{.1135\linewidth}
        \includegraphics[width=1\linewidth]{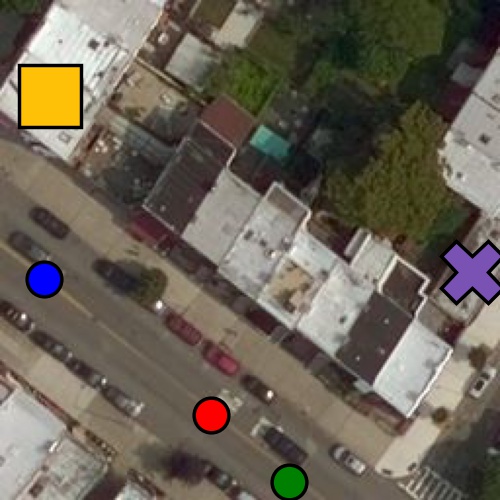}
        \includegraphics[width=1\linewidth]{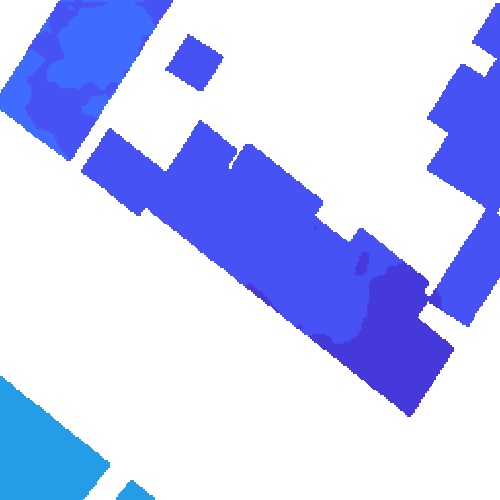}
    \end{subfigure}
    \begin{subfigure}{.29\linewidth}
        \includegraphics[width=1\linewidth]{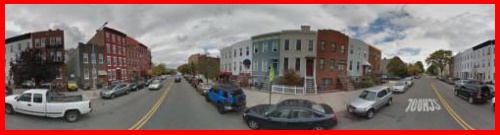}
        \includegraphics[width=1\linewidth]{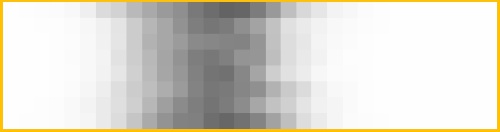}
        \includegraphics[width=1\linewidth]{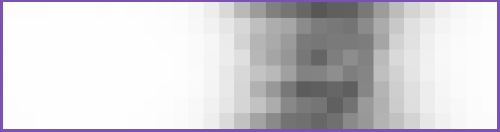}
    \end{subfigure}
    \begin{subfigure}{.29\linewidth}
        \includegraphics[width=1\linewidth]{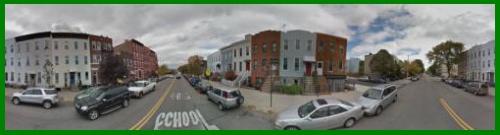}
        \includegraphics[width=1\linewidth]{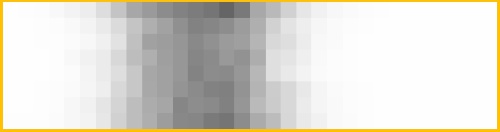}
        \includegraphics[width=1\linewidth]{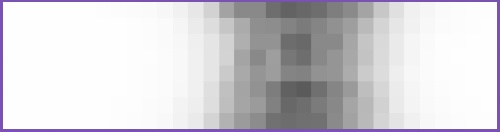}
    \end{subfigure}
    \begin{subfigure}{.29\linewidth}
        \includegraphics[width=1\linewidth]{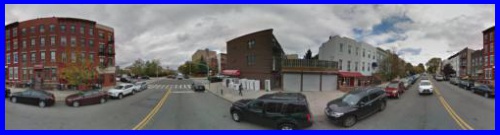}
        \includegraphics[width=1\linewidth]{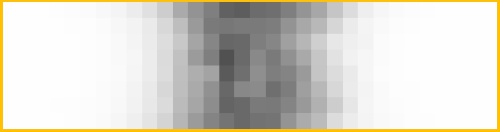}
        \includegraphics[width=1\linewidth]{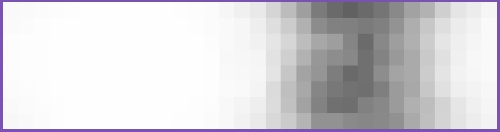}
    \end{subfigure}

  \caption{Spatial attention maps for several ground-level images and target locations using our full method. The location of each panorama is represented by the same-colored dot in the overhead image. For each panorama, the top row of attentions maps corresponds to using the orange $\square$ in the overhead image as the target location, while the bottom row corresponds to using the purple $\times$ as the target location. From top to bottom, the tasks correspond to height estimation, land cover segmentation, and building age prediction.}

  \label{fig:supp_attention_orig}

\end{figure*}

Similarly, \figref{supp_attention_orig} visualizes the spatial attention maps for several pairs of input images and target locations, for three different tasks. For each overhead image, the top row of attention maps corresponds to the $\square$ in the overhead image, and the bottom row corresponds to the $\times$. As expected, the region of high attention is generally oriented toward the target pixel and the attention maps are task dependent. These results demonstrate that our approach is able to learn rich geometric relationships without explicitly providing such supervision and without requiring direct correspondences or other strong geometric assumptions, such as single-image depth estimation.

\section{Extended Evaluation on Queens}

Following the standard protocol, all our models are trained exclusively on the training subset of the Brooklyn portion of the ``Brooklyn and Queens'' dataset~\cite{workman2017unified} (aside from pre-training). In the main paper, we presented results on the held-out testing subset of the Brooklyn portion of the dataset. Here we extend this analysis to show how the model generalizes to the Queens portion. This benchmark is known to be challenging due to large differences in the underlying label distributions and building appearance between the two portions. 

\tabref{queens_results} shows the results of our approach versus baselines on Queens. Our approach, which integrates geospatial attention, generally matches or outperforms two prior methods as well as the single-modality baselines. While there is clearly work left to be done to improve domain adaptation, this result demonstrates that our model is not just over-fitting to the Brooklyn region.

\tabref{brooklyn_ablation_full} extends the ablation study from the main paper, which highlights the importance of the different input features used for geospatial attention, to the remaining tasks (building age, building function, land cover and height). As before, our full approach outperforms baselines, with the geometric features being essential for achieving good performance.

\section{Detailed Architecture}

We provide detailed architecture descriptions for the components of our network. \tabref{arch_overhead} and \tabref{arch_ground} show the feature encoders used for the overhead (EfficientNet-B4) and ground-level (ResNet-50) imagery, respectively. \tabref{arch_grid} shows the architecture for forming the dense ground-level feature map using geospatial attention. \tabref{arch_fusion} corresponds to the fusion network for combining the overhead feature with the dense ground-level feature map. Finally, \tabref{arch_decoder} shows our U-Net style decoder used for generating the segmentation output.

\section{Computational Analysis}

While our method offers significantly improved metrics over the overhead-image only method, it comes at an increase in computational cost. This difference is especially pronounced during training, where a single training run for our full method takes around 67 hours but the overhead-only baseline (\textit{remote}) only required around 8 hours. The ground-only baseline (\textit{proximate}) required around 54 hours to train. We conclude that the primary computational increase is due to the inclusion of the ground-level images. However, we did not extensively optimize for training time computational efficiency. While training time is important, inference time is often a much more important factor in remote sensing applications. We found in our unoptimized implementation that our method requires $\mathord{\sim}0.09$ seconds for a single overhead image (and the corresponding ground-level images). This compares to $\mathord{\sim}0.03$ seconds for the overhead-only baseline.

\clearpage

\begin{table*}
  \centering
  
  \caption{Queens evaluation results.}
  
  \resizebox{\linewidth}{!}{
      \begin{tabular}{@{}lrrrrrrrrcc@{}}
        \toprule
        & \multicolumn{2}{c}{Land Use} & \multicolumn{2}{c}{Age} & \multicolumn{2}{c}{Function} & \multicolumn{2}{c}{Land Cover} & \multicolumn{2}{c}{Height} \\
        & \multicolumn{1}{c}{mIOU} & \multicolumn{1}{c}{Acc} & \multicolumn{1}{c}{mIOU} & \multicolumn{1}{c}{Acc} & \multicolumn{1}{c}{mIOU} & \multicolumn{1}{c}{Acc} & \multicolumn{1}{c}{mIOU} & \multicolumn{1}{c}{Acc} & \multicolumn{1}{c}{RMSE} & \multicolumn{1}{c}{RMSE log} \\
        \bottomrule
        {\em Workman et al.~\cite{workman2017unified}} & 33.48\% & 70.55\% & 9.53\% & 29.76\% & 3.73\% & \textbf{34.13\%} & & & & \\
        {\em Cao et al.~\cite{cao2018integrating}}     & 39.40\% & 74.87\% & & & & & & & & \\
        \hline
        {\em proximate} & 33.84\% & 68.88\% & 10.44\% & 30.13\% & 3.63\% & 33.27\% & 30.02\% & 59.97\% & 4.597 & 1.236 \\
        {\em remote} & 34.16\% & 72.30\% & 8.31\% & 22.91\% & 2.85\% & 29.46\% & \textbf{62.63\%} & 83.54\% & 3.319 & 0.988 \\
        {\em ours} & \textbf{42.93\%} & \textbf{76.85\%} & \textbf{12.88\%} & \textbf{32.93\%} & \textbf{4.08\%} & 34.04\% & 61.24\% & \textbf{83.82\%} & \textbf{3.003} & \textbf{0.946} \\
        \bottomrule
      \end{tabular}
  }
  \label{tbl:queens_results}
\end{table*}

\begin{table*}
  \centering
  
  \caption{Ablation study highlighting the importance of different input features for geospatial attention.}
  
  \resizebox{\linewidth}{!}{
      \begin{tabular}{@{}cccrrrrrrrrrr@{}}
        \toprule
        & & & \multicolumn{2}{c}{Age} & \multicolumn{2}{c}{Function} & \multicolumn{2}{c}{Land Cover} & \multicolumn{2}{c}{Height} \\
        Panorama & Overhead & Geometry & \multicolumn{1}{c}{mIOU} & \multicolumn{1}{c}{Acc} & \multicolumn{1}{c}{mIOU} & \multicolumn{1}{c}{Acc} & \multicolumn{1}{c}{mIOU} & \multicolumn{1}{c}{Acc} & \multicolumn{1}{c}{RMSE} & \multicolumn{1}{c}{RMSE log} \\
        \bottomrule
        \checkmark & &             & 33.52\% & 54.47\% & 13.60\% & 46.53\% & 72.95\% & 87.49\% & 3.128 & 0.766 \\
        & \checkmark &             & 32.49\% & 53.15\% & 14.11\% & 46.44\% & 73.41\% & 87.57\% & 3.135 & 0.781 \\
        & & $d$                    & 39.28\% & 60.02\% & 17.58\% & 51.40\% & 73.83\% & 87.56\% & 3.001 & 0.755 \\
        & & $\theta$               & 37.40\% & 58.57\% & 17.74\% & 50.46\% & 72.90\% & 87.61\% & 3.041 & 0.762 \\
        & & $d, \theta$            & 51.07\% & 70.04\% & 24.21\% & 59.07\% & 72.61\% & 87.93\% & 2.878 & 0.747 \\
        \hline
        \checkmark & \checkmark & $d, \theta$  & \textbf{51.70\%} & \textbf{70.34\%} & \textbf{27.40\%} & \textbf{60.31\%} & \textbf{74.59\%} & \textbf{88.10\%} & \textbf{2.845} & \textbf{0.747} \\
        \bottomrule
      \end{tabular}
  }
  \label{tbl:brooklyn_ablation_full}
\end{table*}

\begin{table*}
  \centering
  \caption{Overhead image feature extractor.}
  
  \begin{tabular}{@{}lcccc@{}}
    \toprule
    Layer (type:depth-idx) & Input Shape & Kernel Shape & Output Shape & Param \# \\
    \bottomrule
    EfficientNet: 1 & -- & -- & -- & -- \\
       — ModuleList: 2-1 & -- & -- & -- & -- \\
       — Conv2dStaticSamePadding: 2-2 & [1, 3, 256, 256] & [3, 48, 3, 3] & [1, 48, 128, 128] & -- \\
       —    — ZeroPad2d: 3-1 & [1, 3, 256, 256] & -- & [1, 3, 257, 257] & -- \\
       — BatchNorm2d: 2-3 & [1, 48, 128, 128] & [48] & [1, 48, 128, 128] & 96 \\
       — MemoryEfficientSwish: 2-4 & [1, 48, 128, 128] & -- & [1, 48, 128, 128] & -- \\
       — ModuleList: 2-1 & -- & -- & -- & -- \\
       —    — MBConvBlock: 3-2 & [1, 48, 128, 128] & -- & [1, 24, 128, 128] & 2,940 \\
       —    — MBConvBlock: 3-3 & [1, 24, 128, 128] & -- & [1, 24, 128, 128] & 1,206 \\
       —    — MBConvBlock: 3-4 & [1, 24, 128, 128] & -- & [1, 32, 64, 64] & 11,878 \\
       —    — MBConvBlock: 3-5 & [1, 32, 64, 64] & -- & [1, 32, 64, 64] & 18,120 \\
       —    — MBConvBlock: 3-6 & [1, 32, 64, 64] & -- & [1, 32, 64, 64] & 18,120 \\
       —    — MBConvBlock: 3-7 & [1, 32, 64, 64] & -- & [1, 32, 64, 64] & 18,120 \\
       —    — MBConvBlock: 3-8 & [1, 32, 64, 64] & -- & [1, 56, 32, 32] & 25,848 \\
       —    — MBConvBlock: 3-9 & [1, 56, 32, 32] & -- & [1, 56, 32, 32] & 57,246 \\
       —    — MBConvBlock: 3-10 & [1, 56, 32, 32] & -- & [1, 56, 32, 32] & 57,246 \\
       —    — MBConvBlock: 3-11 & [1, 56, 32, 32] & -- & [1, 56, 32, 32] & 57,246 \\
       —    — MBConvBlock: 3-12 & [1, 56, 32, 32] & -- & [1, 112, 16, 16] & 70,798 \\
       —    — MBConvBlock: 3-13 & [1, 112, 16, 16] & -- & [1, 112, 16, 16] & 197,820 \\
       —    — MBConvBlock: 3-14 & [1, 112, 16, 16] & -- & [1, 112, 16, 16] & 197,820 \\
       —    — MBConvBlock: 3-15 & [1, 112, 16, 16] & -- & [1, 112, 16, 16] & 197,820 \\
       —    — MBConvBlock: 3-16 & [1, 112, 16, 16] & -- & [1, 112, 16, 16] & 197,820 \\
       —    — MBConvBlock: 3-17 & [1, 112, 16, 16] & -- & [1, 112, 16, 16] & 197,820 \\
       —    — MBConvBlock: 3-18 & [1, 112, 16, 16] & -- & [1, 160, 16, 16] & 240,924 \\
       —    — MBConvBlock: 3-19 & [1, 160, 16, 16] & -- & [1, 160, 16, 16] & 413,160 \\
       —    — MBConvBlock: 3-20 & [1, 160, 16, 16] & -- & [1, 160, 16, 16] & 413,160 \\
       —    — MBConvBlock: 3-21 & [1, 160, 16, 16] & -- & [1, 160, 16, 16] & 413,160 \\
       —    — MBConvBlock: 3-22 & [1, 160, 16, 16] & -- & [1, 160, 16, 16] & 413,160 \\
       —    — MBConvBlock: 3-23 & [1, 160, 16, 16] & -- & [1, 160, 16, 16] & 413,160 \\
       —    — MBConvBlock: 3-24 & [1, 160, 16, 16] & -- & [1, 272, 8, 8] & 520,904 \\
       —    — MBConvBlock: 3-25 & [1, 272, 8, 8] & -- & [1, 272, 8, 8] & 1,159,332 \\
       —    — MBConvBlock: 3-26 & [1, 272, 8, 8] & -- & [1, 272, 8, 8] & 1,159,332 \\
       —    — MBConvBlock: 3-27 & [1, 272, 8, 8] & -- & [1, 272, 8, 8] & 1,159,332 \\
       —    — MBConvBlock: 3-28 & [1, 272, 8, 8] & -- & [1, 272, 8, 8] & 1,159,332 \\
       —    — MBConvBlock: 3-29 & [1, 272, 8, 8] & -- & [1, 272, 8, 8] & 1,159,332 \\
       —    — MBConvBlock: 3-30 & [1, 272, 8, 8] & -- & [1, 272, 8, 8] & 1,159,332 \\
       —    — MBConvBlock: 3-31 & [1, 272, 8, 8] & -- & [1, 272, 8, 8] & 1,159,332 \\
       —    — MBConvBlock: 3-32 & [1, 272, 8, 8] & -- & [1, 448, 8, 8] & 1,420,804 \\
       —    — MBConvBlock: 3-33 & [1, 448, 8, 8] & -- & [1, 448, 8, 8] & 3,049,200 \\
       — Conv2dStaticSamePadding: 2-5 & [1, 448, 8, 8] & [448, 1792, 1, 1] & [1, 1792, 8, 8] & -- \\
       —    — Identity: 3-34 & [1, 448, 8, 8] & -- & [1, 448, 8, 8] & -- \\
       — BatchNorm2d: 2-6 & [1, 1792, 8, 8] & [1792] & [1, 1792, 8, 8] & 3,584 \\
       — MemoryEfficientSwish: 2-7 & [1, 1792, 8, 8] & -- & [1, 1792, 8, 8] & -- \\
    \bottomrule
  \end{tabular}
  
  \label{tbl:arch_overhead}
\end{table*}

\begin{table*}
  \centering
  \caption{Ground-level image feature extractor.}
  
  \begin{tabular}{@{}lcccc@{}}
    \toprule
    Layer (type:depth-idx) & Input Shape & Kernel Shape & Output Shape & Param \# \\
    \bottomrule
    Sequential: 1-1 & [20, 3, 128, 500] & -- & [20, 1024, 8, 32] & -- \\
       — Conv2d: 2-1 & [20, 3, 128, 500] & [3, 64, 7, 7] & [20, 64, 64, 250] & (9,408) \\
       — BatchNorm2d: 2-2 & [20, 64, 64, 250] & [64] & [20, 64, 64, 250] & (128) \\
       — ReLU: 2-3 & [20, 64, 64, 250] & -- & [20, 64, 64, 250] & -- \\
       — MaxPool2d: 2-4 & [20, 64, 64, 250] & -- & [20, 64, 32, 125] & -- \\
       — Sequential: 2-5 & [20, 64, 32, 125] & -- & [20, 256, 32, 125] & -- \\
       —    — Bottleneck: 3-1 & [20, 64, 32, 125] & -- & [20, 256, 32, 125] & (75,008) \\
       —    — Bottleneck: 3-2 & [20, 256, 32, 125] & -- & [20, 256, 32, 125] & (70,400) \\
       —    — Bottleneck: 3-3 & [20, 256, 32, 125] & -- & [20, 256, 32, 125] & (70,400) \\
       — Sequential: 2-6 & [20, 256, 32, 125] & -- & [20, 512, 16, 63] & -- \\
       —    — Bottleneck: 3-4 & [20, 256, 32, 125] & -- & [20, 512, 16, 63] & (379,392) \\
       —    — Bottleneck: 3-5 & [20, 512, 16, 63] & -- & [20, 512, 16, 63] & (280,064) \\
       —    — Bottleneck: 3-6 & [20, 512, 16, 63] & -- & [20, 512, 16, 63] & (280,064) \\
       —    — Bottleneck: 3-7 & [20, 512, 16, 63] & -- & [20, 512, 16, 63] & (280,064) \\
       — Sequential: 2-7 & [20, 512, 16, 63] & -- & [20, 1024, 8, 32] & -- \\
       —    — Bottleneck: 3-8 & [20, 512, 16, 63] & -- & [20, 1024, 8, 32] & 1,512,448 \\
       —    — Bottleneck: 3-9 & [20, 1024, 8, 32] & -- & [20, 1024, 8, 32] & 1,117,184 \\
       —    — Bottleneck: 3-10 & [20, 1024, 8, 32] & -- & [20, 1024, 8, 32] & 1,117,184 \\
       —    — Bottleneck: 3-11 & [20, 1024, 8, 32] & -- & [20, 1024, 8, 32] & 1,117,184 \\
       —    — Bottleneck: 3-12 & [20, 1024, 8, 32] & -- & [20, 1024, 8, 32] & 1,117,184 \\
       —    — Bottleneck: 3-13 & [20, 1024, 8, 32] & -- & [20, 1024, 8, 32] & 1,117,184 \\
    Conv2d: 1-2 & [20, 1024, 8, 32] & [1024, 128, 1, 1] & [20, 128, 8, 32] & 131,200 \\
    LayerNorm: 1-3 & [20, 128, 8, 32] & [8, 128, 32] & [20, 128, 8, 32] & 65,536 \\
    \bottomrule
  \end{tabular}
  \label{tbl:arch_ground}
\end{table*}

\begin{table*}
  \centering
  \caption{Grid architecture.}
  
  \begin{tabular}{@{}lcccc@{}}
    \toprule
    Layer (type:depth-idx) & Input Shape & Kernel Shape & Output Shape & Param \# \\
    \bottomrule
    Grid: 1-1 & -- & -- & [1, 128, 32, 32] & -- \\
    — GeoAttention: 2-1 & -- & -- & [1024, 20, 8, 32] & -- \\
    —    — Conv2d: 3-1 & [20480, 8, 8, 32] & [8, 1, 3, 3] & [20480, 1, 8, 32] & 73 \\
    —    — Conv2d: 3-2 & [20480, 8, 8, 32] & [8, 1, 5, 5] & [20480, 1, 8, 32] & 201 \\
    —    — Conv2d: 3-3 & [20480, 2, 8, 32] & [2, 1, 1, 1] & [20480, 1, 8, 32] & 3 \\
    —    —Sigmoid: 3-4 & [1024, 20, 8, 32] & -- & [1024, 20, 8, 32] & -- \\
    BatchNorm2d: 1-2 & [1, 128, 32, 32] & [128] & [1, 128, 32, 32] & 256 \\
  \end{tabular}
  \label{tbl:arch_grid}
\end{table*}

\begin{table*}
  \centering
  \caption{Fusion (dense ground-level/overhead feature map) architecture.}
  
  \begin{tabular}{@{}lcccc@{}}
    \toprule
    Layer (type:depth-idx) & Input Shape & Kernel Shape & Output Shape & Param \# \\
    \bottomrule
    Conv2d: 1-1 & [1, 184, 32, 32] & [184, 160, 3, 3] & [1, 160, 32, 32] & 265,120 \\
    BatchNorm2d: 1-2 & [1, 160, 32, 32] & [160] & [1, 160, 32, 32] & 320 \\
    ReLU: 1-3 & [1, 160, 32, 32] & -- & [1, 160, 32, 32] & -- \\
    Conv2d: 1-4 & [1, 160, 32, 32] & [160, 160, 3, 3] & [1, 160, 32, 32] & 230,560 \\
    BatchNorm2d: 1-5 & [1, 160, 32, 32] & [160] & [1, 160, 32, 32] & 320 \\
    ReLU: 1-6 & [1, 160, 32, 32] & -- & [1, 160, 32, 32] & -- \\
    Conv2d: 1-7 & [1, 160, 32, 32] & [160, 160, 3, 3] & [1, 160, 32, 32] & 230,560 \\
    BatchNorm2d: 1-8 & [1, 160, 32, 32] & [160] & [1, 160, 32, 32] & 320 \\
    ReLU: 1-9 & [1, 160, 32, 32] & -- & [1, 160, 32, 32] & -- \\
    MaxPool2d: 1-10 & [1, 160, 32, 32] & -- & [1, 160, 16, 16] & -- \\
    Conv2d: 1-11 & [1, 160, 16, 16] & [160, 448, 3, 3] & [1, 448, 16, 16] & 645,568 \\
    BatchNorm2d: 1-12 & [1, 448, 16, 16] & [448] & [1, 448, 16, 16] & 896 \\
    ReLU: 1-13 & [1, 448, 16, 16] & -- & [1, 448, 16, 16] & -- \\
    Conv2d: 1-14 & [1, 448, 16, 16] & [448, 448, 3, 3] & [1, 448, 16, 16] & 1,806,784 \\
    BatchNorm2d: 1-15 & [1, 448, 16, 16] & [448] & [1, 448, 16, 16] & 896 \\
    ReLU: 1-16 & [1, 448, 16, 16] & -- & [1, 448, 16, 16] & -- \\
    Conv2d: 1-17 & [1, 448, 16, 16] & [448, 448, 3, 3] & [1, 448, 16, 16] & 1,806,784 \\
    BatchNorm2d: 1-18 & [1, 448, 16, 16] & [448] & [1, 448, 16, 16] & 896 \\
    ReLU: 1-19 & [1, 448, 16, 16] & -- & [1, 448, 16, 16] & -- \\
    MaxPool2d: 1-20 & [1, 448, 16, 16] & -- & [1, 448, 8, 8] & -- \\
    \bottomrule
  \end{tabular}
  \label{tbl:arch_fusion}
\end{table*}

\begin{table*}
  \centering
  \caption{Decoder architecture.}
  
  \begin{tabular}{@{}lccccc@{}}
    \toprule
    Layer (type:depth-idx) & Input Shape & Kernel Shape &Output Shape & Param \# \\
    \bottomrule
    Upsample: 1-1 & [1, 448, 8, 8] & -- & [1, 448, 16, 16] & -- \\
    DoubleConv: 1-2 & [1, 448, 16, 16] & -- & [1, 448, 16, 16] & -- \\
       — Sequential: 2-1 & [1, 448, 16, 16] & -- & [1, 448, 16, 16] & -- \\
       —    — Conv2d: 3-1 & [1, 448, 16, 16] & [448, 448, 3, 3] & [1, 448, 16, 16] & 1,806,784 \\
       —    — BatchNorm2d: 3-2 & [1, 448, 16, 16] & [448] & [1, 448, 16, 16] & 896 \\
       —    — ReLU: 3-3 & [1, 448, 16, 16] & -- & [1, 448, 16, 16] & -- \\
       —    — Conv2d: 3-4 & [1, 448, 16, 16] & [448, 448, 3, 3] & [1, 448, 16, 16] & 1,806,784 \\
       —    — BatchNorm2d: 3-5 & [1, 448, 16, 16] & [448] & [1, 448, 16, 16] & 896 \\
       —    — ReLU: 3-6 & [1, 448, 16, 16] & -- & [1, 448, 16, 16] & -- \\
    Upsample: 1-3 & [1, 608, 16, 16] & -- & [1, 608, 32, 32] & -- \\
    DoubleConv: 1-4 & [1, 608, 32, 32] & -- & [1, 160, 32, 32] & -- \\
       — Sequential: 2-2 & [1, 608, 32, 32] & -- & [1, 160, 32, 32] & -- \\
       —    — Conv2d: 3-7 & [1, 608, 32, 32] & [608, 160, 3, 3] & [1, 160, 32, 32] & 875,680 \\
       —    — BatchNorm2d: 3-8 & [1, 160, 32, 32] & [160] & [1, 160, 32, 32] & 320 \\
       —    — ReLU: 3-9 & [1, 160, 32, 32] & -- & [1, 160, 32, 32] & -- \\
       —    — Conv2d: 3-10 & [1, 160, 32, 32] & [160, 160, 3, 3] & [1, 160, 32, 32] & 230,560 \\
       —    — BatchNorm2d: 3-11 & [1, 160, 32, 32] & [160] & [1, 160, 32, 32] & 320 \\
       —    — ReLU: 3-12 & [1, 160, 32, 32] & -- & [1, 160, 32, 32] & -- \\
    Upsample: 1-5 & [1, 216, 32, 32] & -- & [1, 216, 64, 64] & -- \\
    DoubleConv: 1-6 & [1, 216, 64, 64] & -- & [1, 56, 64, 64] & -- \\
       — Sequential: 2-3 & [1, 216, 64, 64] & -- & [1, 56, 64, 64] & -- \\
       —    — Conv2d: 3-13 & [1, 216, 64, 64] & [216, 56, 3, 3] & [1, 56, 64, 64] & 108,920 \\
       —    — BatchNorm2d: 3-14 & [1, 56, 64, 64] & [56] & [1, 56, 64, 64] & 112 \\
       —    — ReLU: 3-15 & [1, 56, 64, 64] & -- & [1, 56, 64, 64] & -- \\
       —    — Conv2d: 3-16 & [1, 56, 64, 64] & [56, 56, 3, 3] & [1, 56, 64, 64] & 28,280 \\
       —    — BatchNorm2d: 3-17 & [1, 56, 64, 64] & [56] & [1, 56, 64, 64] & 112 \\
       —    — ReLU: 3-18 & [1, 56, 64, 64] & -- & [1, 56, 64, 64] & -- \\
    Upsample: 1-7 & [1, 88, 64, 64] & -- & [1, 88, 128, 128] & -- \\
    DoubleConv: 1-8 & [1, 88, 128, 128] & -- & [1, 88, 128, 128] & -- \\
       — Sequential: 2-4 & [1, 88, 128, 128] & -- & [1, 88, 128, 128] & -- \\
       —    — Conv2d: 3-19 & [1, 88, 128, 128] & [88, 88, 3, 3] & [1, 88, 128, 128] & 69,784 \\
       —    — BatchNorm2d: 3-20 & [1, 88, 128, 128] & [88] & [1, 88, 128, 128] & 176 \\
       —    — ReLU: 3-21 & [1, 88, 128, 128] & -- & [1, 88, 128, 128] & -- \\
       —    — Conv2d: 3-22 & [1, 88, 128, 128] & [88, 88, 3, 3] & [1, 88, 128, 128] & 69,784 \\
       —    — BatchNorm2d: 3-23 & [1, 88, 128, 128] & [88] & [1, 88, 128, 128] & 176 \\
       —    — ReLU: 3-24 & [1, 88, 128, 128] & -- & [1, 88, 128, 128] & -- \\
    Upsample: 1-9 & [1, 88, 128, 128] & -- & [1, 88, 256, 256] & -- \\
    DoubleConv: 1-10 & [1, 88, 256, 256] & -- & [1, 88, 256, 256] & -- \\
       — Sequential: 2-5 & [1, 88, 256, 256] & -- & [1, 88, 256, 256] & -- \\
       —    — Conv2d: 3-25 & [1, 88, 256, 256] & [88, 88, 3, 3] & [1, 88, 256, 256] & 69,784 \\
       —    — BatchNorm2d: 3-26 & [1, 88, 256, 256] & [88] & [1, 88, 256, 256] & 176 \\
       —    — ReLU: 3-27 & [1, 88, 256, 256] & -- & [1, 88, 256, 256] & -- \\
       —    — Conv2d: 3-28 & [1, 88, 256, 256] & [88, 88, 3, 3] & [1, 88, 256, 256] & 69,784 \\
       —    — BatchNorm2d: 3-29 & [1, 88, 256, 256] & [88] & [1, 88, 256, 256] & 176 \\
       —    — ReLU: 3-30 & [1, 88, 256, 256] & -- & [1, 88, 256, 256] & -- \\
    Conv2d: 1-11 & [1, 88, 256, 256] & [88, 13, 1, 1] & [1, 13, 256, 256] & 1,157 \\
    \bottomrule
  \end{tabular}
  \label{tbl:arch_decoder}
\end{table*}

\end{document}